\documentclass[sigconf, authorversion]{acmart}  %生成适合作者发布的作品版本

\AtBeginDocument{}
\setcopyright{acmcopyright}
\copyrightyear{2023}
\acmYear{2023}
\acmDOI{XXXXXXX.XXXXXXX}

\acmConference[MM '23]{Proceedings of the 31st ACM International Conference on Multimedia}{October 28, 2023 -- November 3, 2023}{Ottawa, Canada}
\acmBooktitle{Proceedings of the 31st ACM International Conference on Multimedia (MM '23), October 28, 2023 -- November 3, 2023, Ottawa, Canada}

\acmPrice{15.00}
\acmISBN{978-1-4503-XXXX-X/18/06}

\acmSubmissionID{533}

\usepackage{algorithm}
\usepackage{algpseudocode}  
\usepackage{amsmath}  
\usepackage{color}
\usepackage{bm}
\usepackage{multirow}
\usepackage{wrapfig}
\usepackage{subcaption}
\usepackage{booktabs}

\newcommand{\tr}{{\rm Tr}}

\newcommand{\eg}{\textrm{e.g.}}
\newcommand{\ie}{\textrm{i.e.}}

\renewcommand{\leq}{\leqslant}

\newtheorem{defi}{Definition}

\begin{document}

\title{Self-Contrastive Graph Diffusion Network}

\author{Yixuan Ma}
\email{mayx2021@lzu.edu.cn}
\affiliation{%
  \institution{School of Information Science and Engineering, \\
  	Lanzhou University}
  \city{Lanzhou}
  \country{China}
}

\author{Kun Zhan}
\authornote{Corresponding author.}
\email{kzhan@lzu.edu.cn}	
\affiliation{%
\institution{School of Information Science and Engineering, \\
	Lanzhou University}
\city{Lanzhou}
\country{China}
}

\renewcommand{\shortauthors}{Ma and Zhan}

\begin{abstract}
Contrastive learning has been proven to be a successful approach in graph self-supervised learning. Augmentation techniques and sampling strategies are crucial in contrastive learning, but in most existing works, augmentation techniques require careful design, and their sampling strategies can only capture a small amount of intrinsic supervision information. Additionally, the existing methods require complex designs to obtain two different representations of the data. To overcome these limitations, we propose a novel framework called the Self-Contrastive Graph Diffusion Network (SCGDN). Our framework consists of two main components: the Attentional Module (AttM) and the Diffusion Module (DiFM). AttM aggregates higher-order structure and feature information to get an excellent embedding, while DiFM balances the state of each node in the graph through Laplacian diffusion learning and allows the cooperative evolution of adjacency and feature information in the graph. Unlike existing methodologies, SCGDN is an augmentation-free approach that avoids "sampling bias" and semantic drift, without the need for pre-training.  We conduct a high-quality sampling of samples based on structure and feature information. If two nodes are neighbors, they are considered positive samples of each other. If two disconnected nodes are also unrelated on $k$NN graph, they are considered negative samples for each other. The contrastive objective reasonably uses our proposed sampling strategies, and the redundancy reduction term minimizes redundant information in the embedding and can well retain more discriminative information. In this novel framework, the graph self-contrastive learning paradigm gives expression to a powerful force. SCGDN effectively balances between preserving high-order structure information and avoiding overfitting. The results manifest that SCGDN can consistently generate outperformance over both the contrastive methods and the classical methods.
\end{abstract}




\maketitle

%______________________________Introduction_____________________________%
\section{Introduction}

\begin{figure}[!ht]
\centering
	  \includegraphics[width=0.48\textwidth]{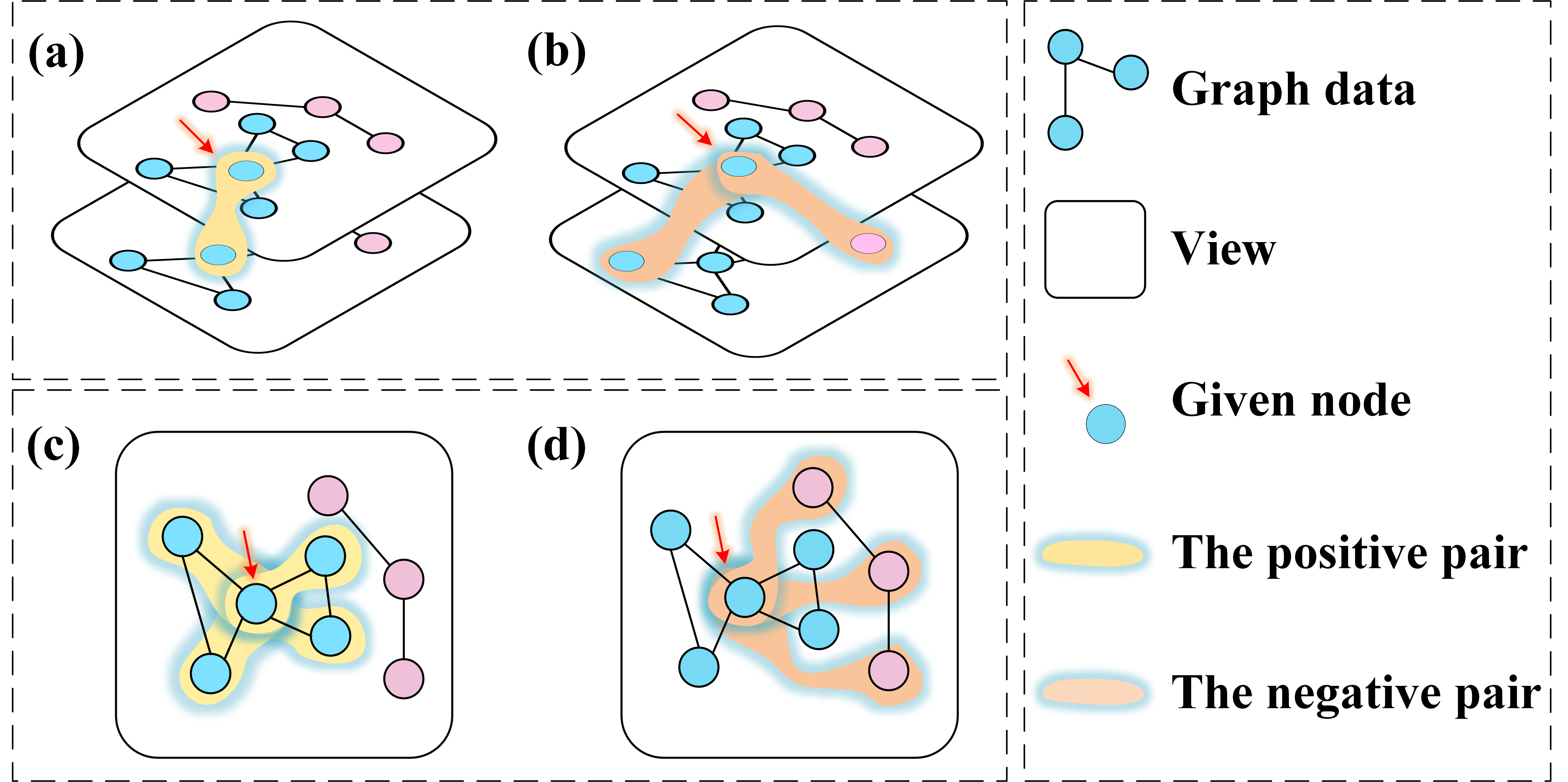}
        \caption{A comparison between contrastive strategies for the prior works and ours. 
Subfigures (a) and (b) depict the common contrastive strategies, where a positive sample pair consists of the same node in two different views, while a negative sample pair is randomly selected. In contrast, subgraphs (c) and (d) demonstrate our proposed self-contrastive strategies, which rely on the intrinsic and valuable structure and feature information of the data for positive and negative samplings.}
  \label{Motivation}
\end{figure}

Deep graph clustering is a fundamental yet hot topic in the graph field, which has attracted much attention for decades. The goal of deep graph clustering is to partition a given graph, where the edges between groups have very low weights and the edges within the group have high weights.  The existing methods of deep graph clustering can be roughly divided into three categories: generative methods~\cite{wang2017mgae,wang2019attributed,cheng2020multiview}, adversarial methods~\cite{zhang2019attributed,pan2019learning,bo2020structural,tu2021deep}, and contrastive methods~\cite{hassani2020contrastive,cui2020adaptive,zhao2021graph,lee2021augmentation,gong2022attributed,xia2022progcl}. 
Our proposed method belongs to the contrastive learning category. 

Inspired by the success of contrastive learning in computer vision (CV)~\cite{oord2018representation,hjelm2018learning}, a growing number of works have been adapted to deep graph clustering~\cite{hassani2020contrastive,pan2021multi, gong2022attributed}. Although contrastive graph clustering has achieved impressive performance, they require complex designs with a mass of parameters. For instance, MVGRL~\cite{hassani2020contrastive} harness two dedicated Graph Neural Networks (GNNs) as graph encoders, a graph pooling layer as the readout function, and a discriminator as a parameterized mutual information estimator.  Similarly, AFGRL~\cite{lee2021augmentation} updates the online encoder parameters to the target encoder parameters using the Exponential Moving Average (EMA) and Stop-Gradient. Graph Convolutional Networks (GCNs) have shown remarkable performance on many network analysis tasks. However, most GCNs methods may not be applicable to real-world scenarios where the graph is dynamic. In contrast, Graph Diffusion Networks (GDNs) aim to capture the dynamic nature of the graph, by iteratively diffusing node features across the graph~\cite{chamberlain2021grand, chamberlain2021blend, chen2022optimization}. Hence, a natural question emerges: \textbf{\textit{how to design a model framework with fewer parameters to contrastive graph clustering? }}
Furthermore, the traditional graph contrastive learning paradigm mainly leverages corruptions and the momentum encoder to construct negative pairs. In fact, the process of constructing negative samples is random. While some studies~\cite{lee2021augmentation, zhao2021graph, yang2023cluster} have recognized the importance of crucial clustering information in enhancing the discriminative capability, many works require pre-training models to obtain more accurate clustering information. This observation motivates us to reconsider that \textbf{\textit{how to utilize intrinsic data information to develop an effective graph contrastive learning paradigm?}}
 
In light of these above issues, we propose a novel approach called the Self-Contrastive Graph Diffusion Network (SCGDN), which is depicted in Fig.~\ref{framework}. Specifically, the architecture of SCGDN consists of two components. The first component, which we refer to as the Attentional Module (AttM), encodes  feature and the high-order structure information of nodes into the latent space to guarantee learning excellent representation. 
There is an intuition that each node in the graph constantly changes its state until the final balance is achieved due to the influence of neighbors and distant points. That means we can utilize the excellent representation for diffusion learning. Therefore, the second component is called the Diffusion Module (DiFM).  In DiFM, diffusion for $t$ time steps acts as a continuous analog of layers to aggregate information from $t$-hop neighbors.  Inheriting the spirit of neural ODEs~\cite{chen2018neural}, the derivative of the hidden layer state parameterized by the neural network. This allows us to design a model framework with fewer parameters and develop more efficient diffusion process. 
The proposed SCGDN employs AttM to preserve the high-order structure information in the original feature space, and applies DiFM to capture the dynamic nature of the graph by diffusing node features in graphs.

To further improve the performance of contrastive graph clustering, we have inherited the advantages of COLES~\cite{zhu2021contrastive} and the success of redundancy reduction in latent space. In the unsupervised setting, contrastive learning requires generating two augment views of the same data samples, in which the same node itself is considered positive (as shown in Fig.~\ref{Motivation} (a)). However, the common negative sampling strategy is to randomly sample another node, treating a given node and another node as negative pairs (as shown in Fig.~\ref{Motivation} (b)).
Under this strategy, there may be a link or feature similarity between the negative pairs, which is contradictory to the "negativeness". 
In fact, the structure and feature properties of graph data are valuable, \textit{\ie}, similar nodes may have link or feature similarity. 
Fig.~\ref{Motivation} (c) and (d) show the motivation that the positive set for a given node is supposed to be a set of nodes that are associated with the given node, and the negative set for a given node should be a set of nodes that are not related in structure or feature to the given node. 
Based on this observation, we further propose to leverage a block loss to construct a contrastive learning objective for learning more effective and abundant supervision information.  
The contrastive objective reasonably uses internal information to sample high-quality positive and negative samples, and the redundancy reduction term minimizes redundant information in the embedding and can well retain more discriminative information. Through their coordinated guidance, the potential spatial quality of subsequent clustering tasks is ensured. Hence, SCGDN cleverly avoids the asymmetric bi-encoders and the Siamese networks using a novel graph contrastive learning paradigm, addressing previous concerns. 

The main contributions are summarized as follows: 
\begin{itemize}
\item We propose a novel Self-Contrastive Graph Diffusion Network (SCGDN), which effectively balances between preserving high-order structure information and avoiding overfitting. 
\item We design an augmentation-free and free pre-training model framework, which avoids the "sampling bias" and the semantic drift while avoiding complex model designs, including two main parts, \textit{\ie},  AttM and DiFM. 
\item We introduce and theoretically analyze a novel graph contrastive learning paradigm that conducts contrastive learning with the proposed high-quality sampling strategies and without multiview. 
\end{itemize}

To sum up, SCGDN offers an exceptional model framework and optimization paradigm that can achieve remarkable clustering performances. We conducted a comprehensive evaluation of SCGDN on six benchmark datasets for graph clustering, and the findings indicate that SCGDN outperforms both the contrastive and classical methods in terms of performance gains. Notably, the results show the effectiveness of the proposed approach and its potential for broader applications in the field of graph clustering.

%______________________________Related Work_____________________________%
\section{Related Work}
The contrastive methods are one of the most powerful methods in self-supervised learning. The goal is to push similar nodes closer, while pulling different nodes farther. In a nutshell, we will describe the following three key issues. 

\textbf{Model framework.}
In graph representation learning, GCN~\cite{kipf2016semi} has become the almost de facto and widely adopted encoder. For example, MVGRL~\cite{hassani2020contrastive} uses un-shared GCNs and a shared MLP. DCRN~\cite{liu2022deep} and GDCL~\cite{zhao2021graph} use the shared parameters Siamese networks. There are a vast number of works~\cite{lee2021augmentation,ijcai2021p204} using other asymmetric bi-encoders, including EMA, momentum update, and Stop-Gradient. Nevertheless, a tremendous number of previous works either builg a parallel framework which has asymmetrical bi-encoders with plenty of parameters or enter two different views into a siamese network with shared parameters. In contrast, our proposed SCGDN only contains an end-to-end graph diffusion network. What's more important, SCGDN has an augmentation-free and free pre-training model framework, which avoids the "sampling bias" and the semantic drift.

\textbf{Sampling strategies.}
The key detail of contrastive methods is how to characterize high-quality positive and negative samples. The outstanding DGI~\cite{velickovic2019deep} in graph contrastive learning, inspired by the prior success of DIM~\cite{hjelm2018learning}, treats each local representation and the summarized graph-level representation as a positive sample pair, while negative sample pairs are defined a shuffle representation and the summarized graph-level representation.
Inspired by DGI~\cite{velickovic2019deep}, MVGRL~\cite{hassani2020contrastive} utilizes diffusion matrices and adjacency matrices as graph structure information, and also uses random shuffle to construct negative samples. GCC~\cite{qiu2020gcc} uses the negative sampling strategy proposed by MOCO~\cite{he2020momentum}.
Later, much works~\cite{zhu2020deep,you2020graph,zhu2021graph} focus on how to construct different types of augmentation views, positive sample pairs, and negative sample pairs.
Until the emergence of AFGRL~\cite{lee2021augmentation} breaks the above situation, which required neither augmentation nor negative sampling. The positive samples of AFGRL~\cite{lee2021augmentation} are determined by the adjacency matrix, the nearest neighbor obtained from learning, and the cluster information.
COLES~\cite{zhu2021contrastive} randomly generates negative samples with a Gaussian distribution based on the random graph sampling theory~\cite{Erdos1959pmd}. Beyond these, we introduce a high-quality negative sampling strategy, which depends on the adjacency matrix and $k$NN graph.

\textbf{Objective function.}
The theoretical core of contrastive learning is the InfoNCE principle~\cite{oord2018representation}, which maximizes the mutual information between different representations. With the introduction of DIM~\cite{hjelm2018learning},
MVGRL~\cite{hassani2020contrastive} which applies InfoMax loss, focuses on maximizing the mutual information between the local representation and the global representation.
To be different, DCRN~\cite{liu2022deep} considers the sample-level and feature-level of correlation reduction and designs the MSE loss to the identity matrix as well as the reconstruction loss and the clustering loss.
In particular, AFGRL~\cite{lee2021augmentation} minimizes the cosine distance between the positive pairs.
The work~\cite{boudiaf2020unifying} has uncovered tight relations between the cross-entropy loss and the contrastive loss, which inspires future studies in the unsupervised learning area.
In addition, COLES~\cite{zhu2021contrastive} reformulates the Laplacian Eigenmaps~\cite{belkin2003laplacian} into contrastive learning. Furthermore, we propose an efficient graph contrastive learning paradigm with the representation level of correlation reduction.

%______________________________Method_____________________________%
\section{Methodology}
\begin{figure*}[!ht]
\centering
	  \includegraphics[width=0.92\textwidth]{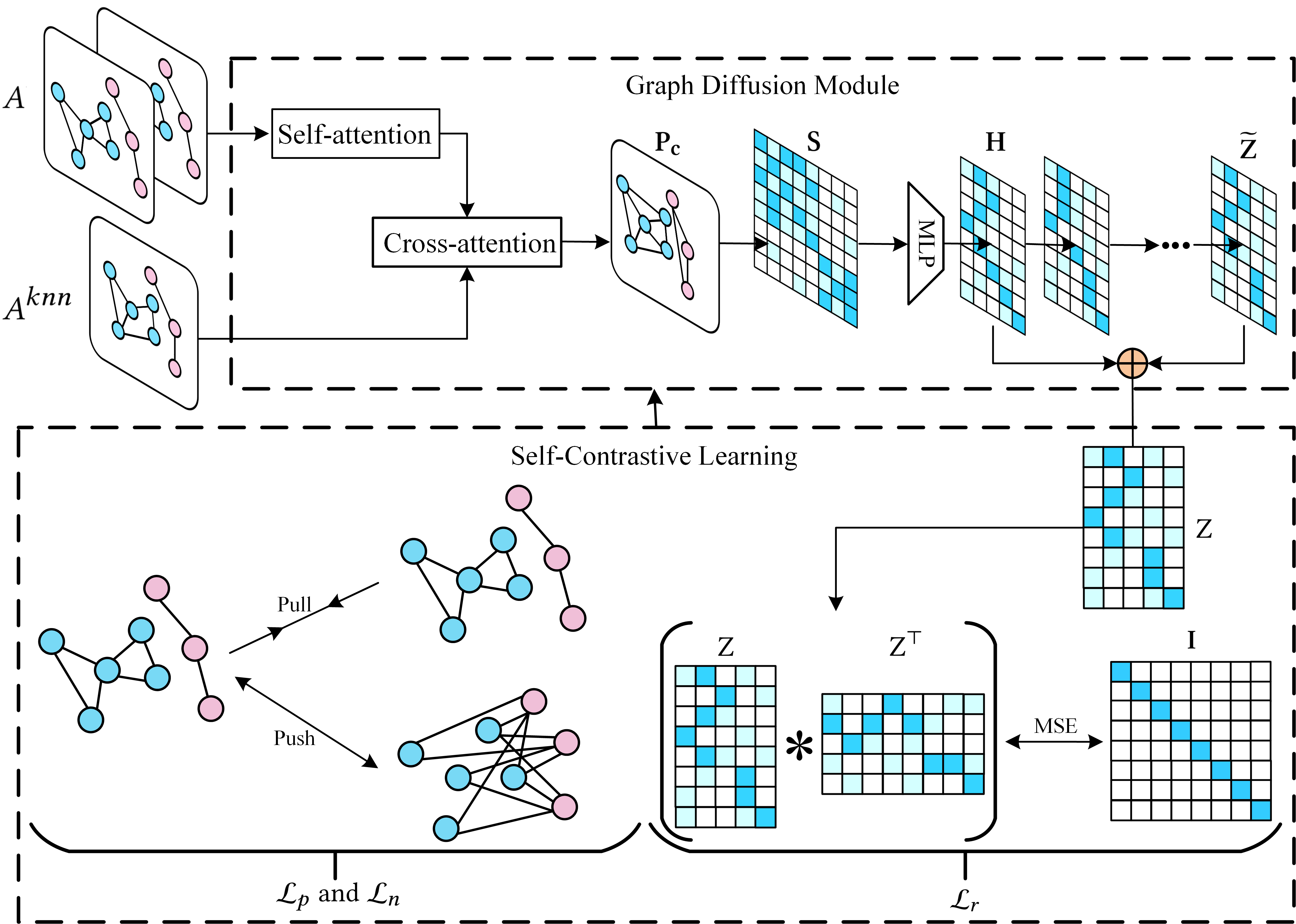}
        \caption{The illustration of the proposed Self-Contrastive Graph Diffusion Network (SCGDN). Given an undirected attribute graph, we first calculate the representation of samples by the graph diffusion module. Then, the Self-Contrastive Learning objective guides the update of the parameters of the graph diffusion module.}
  \label{framework}
\end{figure*}
In this section, we propose a novel Self-Contrastive Graph Diffusion Network (SCGDN). The overall framework of SCGDN is shown in Fig.~\ref{framework}. Then, we introduce the proposed SCGDN in detail from the graph diffusion module and the self-contrastive learning objective. 

\subsection{Notations and Preliminaries}
Given an undirected graph $\mathcal{G}= ( \mathbf{V}, \mathbf{E}, \mathbf{X} )$ , where $\mathbf{V}=[v_{1}, v_{2}, \ldots, v_{n}] \in \mathbb{R}^{n}$ represents $n$ nodes, $\mathbf{X}= [\mathbf{x}_{1}, \mathbf{x}_{2}, \ldots, \mathbf{x}_{n}]^{\top} \in \mathbb{R}^{n \times d}$ is the corresponding feature matrix of the nodes, and $\mathbf{E}$ is a set of edges denoted by an adjacency matrix $\widetilde{\mathbf{W}}=[\widetilde{w}_{i j}] \in \mathbb{R}^{n \times n}$, where $\widetilde{w}_{i j}=1$ if $\left(v_{i}, v_{j}\right) \in \mathbf{E}$ and $\widetilde{w}_{i j}=0$ otherwise.
$\mathbf{W}=\mathbf{D}^{-\frac{1}{2}}(\widetilde{\mathbf{W}}+\mathbf{I}) \mathbf{D}^{-\frac{1}{2}} \in \mathbb{R}^{n \times n}$ is a symmetrically normalized adjacency matrix, $\mathbf{D} \in \mathbb{R}^{n \times n}$ is a diagonal matrix containing degrees of nodes, and $\mathbf{I} \in \mathbb{R}^{n \times n}$ represents the identity matrix.

Before encoding, we use the widely used $k$NN graph $\mathbf{W}^{knn}$ to build feature neighbor information aggregation, which encodes the similarities of each node feature. Unlike other work~\cite{LiuAAAI2022} that utilizes the Gaussian kernel, we use the t-distribution kernel. As is known to all, the t-distribution is more gentle and more robust than the Gaussian distribution.

\subsection{Graph Diffusion Module}
\subsubsection{Attentional Module}
Inspired by the success of SDSNE~\cite{LiuAAAI2022}, its intuition is a multiview system needs to share the intrinsical structure information by a shared self-attentional module.
Inheriting the power of SDSNE~\cite{LiuAAAI2022}, we jointly embed the structure  and feature information of nodes into the latent space by designing a novel Attentional Module (AttN). The proposed AttN contains a self-attentional layer and a cross-attentional layer, respectively. Mathematically,
\begin{align}\label{AttM}
\mathbf{P}_s &= \mathbf{W} \boldsymbol{\Theta}_{1} \mathbf{W}^{\top} , \\
\mathbf{P}_c &= \mathbf{P}_s \boldsymbol{\Theta}_{2} \mathbf{W}^{knn} ,
\end{align}  
where $\mathbf{P}_s, \mathbf{P}_c \in \mathbb{R}^{n \times n}$ denote the attentional graphs, $\boldsymbol{\Theta}_{1}$ and $\boldsymbol{\Theta}_{2}$ are the trainable parameters. It is worth mentioned that the self-attentional layer explores higher-order structural information. Through the cross-attentional layer, feature information is better aggregated. 

Subsequently, we calculate the similarity matrix $\mathbf{S}$ and normalize the similarity matrix $\mathbf{S}$ with $\ell^2$-norm as formulated: 
\begin{align}\label{simi}
\mathbf{S} = \mathbf{P_c} \mathbf{P_c}^{\top}, 
\quad \mathbf{S} =[\textbf{s}_i]\in \mathbb{R}^{n \times n},
\quad \mathbf{s}_i = \frac{\mathbf{s}_i}{\|\mathbf{s}_i\|_2}, \forall\,i,
\end{align}
where $s_i$ denotes a column in $\mathbf{S}$.

Then, we encode the similarity matrix $\mathbf{S}$ with two separated linear layers called MLP as follows:
\begin{align}\label{mlp}
\mathbf{H}= \operatorname{MLP}_{\boldsymbol{\Phi}_1,\boldsymbol{\Phi}_2}(\mathbf{S}) \in \mathbb{R}^{n \times d'},
\end{align}
where $d'$ is the number of hidden dimensions, $\boldsymbol{\Phi}_1$ and $\boldsymbol{\Phi}_2$ are the trainable parameters of linear layers, respectively. 

Through AttM, the higher-order structure and feature information of graph data are better aggregated, thus improving the expression ability of graph representation and further improving the performance of the downstream tasks.

\subsubsection{Diffusion Module}
Recently, some works~\cite{chamberlain2021grand, chamberlain2021blend, chen2022optimization} have solved the common difficulties of graph learning model, such as depth, over-smoothing, etc. 
Motivated by their success, we introduce Diffusion Module (DiFM) to learning node embedding. 

\begin{defi}[Graph diffusion] 
A graph space consists of feature and structure information $\mathbf{Z} = (\mathbf{X}, \mathbf{W})$. For a node, graph diffusion with time-dependent $t$ can be achieved as follows: 
\begin{equation}\label{DiFM1}
\begin{split}
\frac{ \partial \widetilde z_i \left( t \right)}{\partial t} = {\rm div} \left(\mathbf{a}_i \left(\mathbf{z}_i (t) \right) \nabla \widetilde z_i(t) \right), \\
\quad \widetilde z_i(0) = \mathbf{h}_i; \quad i =1,...,n; \quad t \ge 0,
\end{split}
\end{equation}
where the function $\mathbf{a}_i(\cdot)$ is the diffusivity controlling the diffusion strength between node $i$ and its neighbors. 
\end{defi} 

\begin{defi}[The stationary state of graph diffusion] 
In order to produce a stationary state of graph diffusion, DiFM should be able to learn the overall information of the graph. By Eq.~(\ref{DiFM1}), the stationary state $\mathbf{\widetilde{Z}}$  for time $t$ can be written as: 
\begin{align}\label{DiFM2}
\mathbf{\widetilde{Z}}(t) =  \mathbf{\widetilde{Z}}(0) + \int_{0}^{t}\frac{\partial {\mathbf{\widetilde{Z}} (\tau)}}{\partial\tau}{\rm d}\tau,
\end{align}
where $\tau$ is the time step.
\end{defi} 

Next, to obtain the integral term, we must compute $\frac{\partial {\mathbf{\widetilde{Z}} \left(t \right)}}{\partial t}$. Here, we borrow from the Neural ODEs~\cite{chen2018neural}, that is, the derivative of the hidden layer state parameterized by the neural network $f$,
\begin{equation}\label{DiFM_de}
\frac{\partial {\mathbf{\widetilde{Z}} \left(t \right)}}{\partial t} = f\left(\mathbf{\widetilde{Z}}(t),\mathbf{W}, t , \boldsymbol{\varPsi} \right)  = \left(\mathbf{W}-\mathbf{I} \right)\mathbf{\widetilde Z}(t)
\end{equation}
where $\mathbf{W} =[w_{ij}], \forall ij, w_{ij}= w_{ji} \ge 0, \sum_i w_{ij} = 1$, and $\boldsymbol{\varPsi}$ is the trainable parameters of the neural network $f$.

Graph diffusion is interpreted as a nonlinear filter depending on feature information and adjacency relations. In the limit $t \to \infty$, the graph becomes stable and each connected component is equal to its average feature~\cite{LiuAAAI2022}. This emphasizes the observation that adjacent nodes have closer relationships than non-adjacent nodes, and nodes can propagate messages to neighbors, making adjacent nodes closer in terms of feature.

In addition, we further sum the input embedding $\mathbf{H}$ and the stationary state $\mathbf{\widetilde{Z}}$ with the normalization factor $\omega$ as formulated:
\begin{align}\label{DiFM3}
\mathbf{Z}= \sigma \left(\omega \mathbf{\widetilde Z}+\mathbf{H} \right), 
\end{align}
where $\sigma(\cdot)$ is an activation function, \textit{\eg}, ReLU($\cdot$) = max(0;$\cdot$), and the normalization factor $\omega=\sqrt{2 \mathbf {D}}$ utilizes the degree information of each node. Following, we normalize $\mathbf{Z}$ with $z$-score normalization. Finally, we perform K-means on the optimal embedding $\mathbf{Z}$ obtained. 

DiFM has the following advantages. First, by injecting graph relations into feature information, it generates more useful node representations for downstream tasks. Second, it allows the cooperative evolution of adjacency and feature information in a graph.

\subsection{Self-Contrastive Learning}
Driven by the classic InfoNCE loss, plenty of works~\cite{lee2021augmentation,xia2022progcl} have achieved excellent clustering performance. Yet, the classic InfoNCE loss is bounded by JS divergence, which yields zero and vanishing gradients.
Consider the block-contrastive loss COLES~\cite{zhu2021contrastive}, which realizes the negative sampling strategy for Laplacian Eigenmaps, is driven by reformulating SampledNCE into Wasserstein GAN using a GAN-inspired contrastive formulation.
\begin{align}\label{COLES}
\mathcal{L} = \operatorname{Tr}\left(\mathbf{Z}^{\top} \mathbf{L^{(+)}} \mathbf{Z}\right)-\frac{\eta^{\prime}}{\kappa} \sum_{k=1}^{\kappa} \operatorname{Tr}\left(\mathbf{Z}^{\top} \mathbf{L}_{k}^{(-)} \mathbf{Z}\right),
\end{align}
where $\mathbf{L^{(+)}}$ is degree-normalized Laplacian matrix capturing the positive sampling, $\mathbf{L}_{k}^{(-)}$ for $k=1,...,\kappa$ are randomly generated degree-normalized Laplacian matrices capturing the negative sampling, and $\eta^{\prime}$ is a scalar to controlling the effect of negative samples.

However, we find the drawback of COLES~\cite{zhu2021contrastive} is that the negative sample set is randomly generated, which may mislead the model into learning wrong parameters due to the inaccurate negative samples. To solve this problem, we propose a novel negative sampling strategy to get an adjacency matrix of negative samples. 

\subsubsection{Negative Sampling}
Many methods are used to shuffle the index~\cite{velickovic2019deep,hassani2020contrastive} or randomly generate negative samples~\cite{zhu2021contrastive}. However, this would violate the negativity of negative samples.  
In order to generate high-quality negative samples, we consider both structure and feature information. Mathematically, we define an adjacency matrix of negative samples as follows: 
\begin{align}\label{neg}
\widetilde{w}_{ij}^{(-)} =
\begin{cases}
  1, & {\rm if }\, w_{ij} \cup w_{ij}^{\rm knn} = 0 \\
  0, & {\rm otherwise\,. }
\end{cases}
\end{align}
For the convenience of narrative, we denote $\widetilde{\mathbf{W}}^{(+)} = \widetilde{\mathbf{W}}$  and $\widetilde{\mathbf{D}}^{(+)} =\mathbf{D}$ as the adjacency matrix and the diagonal matrix of positive samples, respectively.

\subsubsection{Objective}
Based on the negative sampling, we formulate the objective function as follows:
\begin{equation}\label{loss}
\begin{split}
\mathcal{L} &= \mathcal{L}_{p} - \boldsymbol{\beta} \mathcal{L}_{n}+ \gamma \mathcal{L}_{r} \\
&= \tr \left(\mathbf{Z}^{\top} \mathbf{L}^{(+)} \mathbf{Z}\right) - \beta  \tr \left(\mathbf{Z}^{\top} \mathbf{L}^{(-)} \mathbf{Z}\right) + \gamma \| \mathbf{Z} \mathbf{Z}^{\top}-\mathbf{I}\|_{\rm F}^{2}\,,
\end{split}
\end{equation}
where $\beta$ is a non-negative hyperparameter trading off  the two contrastive terms, $\mathbf{L}^{(+)} = \mathbf{I} - (\widetilde{\mathbf{D}}^{(+)})^{-\frac{1}{2}}\widetilde{\mathbf{W}}^{(+)} (\widetilde{\mathbf{D}}^{(+)})^{-\frac{1}{2}}$ and $\mathbf{L}^{(-)} = \mathbf{I} - (\widetilde{\mathbf{D}}^{(-)})^{-\frac{1}{2}}\widetilde{\mathbf{W}}^{(-)} 
 (\widetilde{\mathbf{D}}^{(-)})^{-\frac{1}{2}}$ are the Laplacian matrix of positive samples and negative samples, respectively. In addition, $\gamma$ is also a non-negative hyperparameter, and it controls the last term which encourages incoherence between the column vectors of network output.  The detailed learning process of SCGDN is shown in Algorithm~\ref{algorithm}.
\begin{algorithm}[h]
\caption{The SCGDN algorithm.}\label{algorithm}
\begin{algorithmic}[1]
\Require
feature matrix $\mathbf{X}$;
 adjacency matrix  $\widetilde{\mathbf{W}}$; 
the number of neighbors $k$; 
and parameters $\beta$, $\gamma$\,.
\Ensure the clustering results $R$\,.
\State \textbf{Initialization}: $epoch=1$, $epoch_{\max}$, and the model parameters\,.
\State Build $k$NN graphs $\mathbf{W}^{knn}$ for the feature matrix $\mathbf{X}$ with the t-distribution kernel\,;
\While{$epoch\leq {epoch_{\max}}$}
    \State Obtain the attentional graphs $\mathbf{P}_s$ and $\mathbf{P}_c$ by Eq. (1) and Eq. (2)\,;
    \State Calculate the similarity matrix $\mathbf{S}$ and normalize the similarity matrix  $\mathbf{S}$  by Eq.~(\ref{simi})\,;
    \State Encode the similarity matrix $\mathbf{S}$ with MLP encoder by Eq.~(\ref{mlp})\,;
    \State Obtain the stationary state  $\widetilde{\mathbf{Z}}$ using the Diffusion Module (DiFM) with Eq.~(\ref{DiFM2})\,;
    \State Calculate the node representation $\mathbf{Z}$ by Eq.~(\ref{DiFM3}) and normalize $\mathbf{Z}$ with z-score normalization\,;
    \State Update parameters by minimizing $\mathcal L$ in Eq.~(\ref{loss})\,;
    \State $epoch = epoch + 1$\,;
\EndWhile
\State Perform K-means on $\mathbf{Z}$ to obtain the final clustering results $R$\,.
\State \Return $R$\,.
\end{algorithmic}
\end{algorithm}

%______________________________Experiments_____________________________%
\section{Experiments}
We conduct various experiments to evaluate the effectiveness and efficiency of the proposed SCGDN method on the node clustering task.
The focuses of the experiments are to validate the representation ability of the learned feature, the effectiveness of the model framework, and the necessity of each component of the objective function.

\textbf{Datasets.\,}
We compare our SCGDN approach to different baselines on six benchmark datasets, including Cora~\cite{kipf2016variational}, Citeseer~\cite{kipf2016variational},  Brazil Air-Traffic (BAT)~\cite{mrabah2022rethinking}, Europe Air-Traffic (EAT)~\cite{mrabah2022rethinking}, CoraFull~\cite{liu2022deep} and  Amazon Photo (AMAP)~\cite{liu2022deep}.
The statistics of these datasets are summarized in Table~\ref*{dataset}, and the descriptions are as follows.
\begin{itemize}
\item Cora~\cite{kipf2016variational}, Citeseer~\cite{kipf2016variational} and CoraFull~\cite{liu2022deep} are well-known citation network datasets. 
Nodes represent papers, and edges indicate the citation relationship. 
The bag-of-words representation of papers are regarded as node features, and labels are academic fields.

\item BAT~\cite{mrabah2022rethinking} and EAT~\cite{mrabah2022rethinking} are two air-traffic datasets (Brazil and Europe). 
Nodes correspond to airports, and edges indicate the existence of commercial flights. 
Node features are constructed by leveraging the one-hot encoding of node degrees.
Labels are corresponding to the airport's level of activity, measured in flights or people.

\item AMAP~\cite{liu2022deep} is based on Amazon's co-purchase data. Nodes denote products, while edges reflect the two products are purchased at the same time. There are the sparse bag-of-words attribute vector encoding product reviews as node features, and labels are product categories.
\end{itemize}

\begin{table}[!ht]
  \centering
  \caption{Statistics summary of the graph datasets.}
  \label{dataset}
  \centering
  	\begin{tabular*}{\linewidth}{@{\extracolsep{\fill}\,}lcccc}
 	 \toprule
		Dataset &Nodes &Edges &Features  &Clusters  \\
	 	 \midrule  % hline
		Cora &2,708 & 5,429 & 1,433 & 7  \\
		Citeseer  &3,327 & 4,732 & 3,703 & 6  \\
		AMAP & 7,650 & 119,081 & 745 & 8 \\
		BAT &131 &1,038 &81 &4  \\
		EAT &399 &5,994 &203 &4  \\
		CoraFull &19,793 &63,421 &8,710 &70   \\
    \bottomrule
    \end{tabular*}
\end{table}

\textbf{Baseline models.\,}
To compare SCGDN with previous works, we choose three types of deep clustering methods as baselines, including generative methods (GAE~\cite{kipf2016variational}, MGAE~\cite{wang2017mgae}, DAEGC~\cite{wang2019attributed}), adversarial methods (DFCN~\cite{tu2021deep}, SDCN~\cite{bo2020structural}), and contrastive methods (MVGRL~\cite{hassani2020contrastive}, DCRN~\cite{liu2022deep}, GDCL~\cite{zhao2021graph}, AutoSSL~\cite{jin2021automated}, AGC-DRR~\cite{gong2022attributed}, AFGRL~\cite{lee2021augmentation}, ProGCL~\cite{xia2022progcl}). 
For the results of all data from the baselines, we will quote the results directly or quote the results of the baseline being replicated.

\begin{table*}[!ht]\setlength{\tabcolsep}{1.5pt}
    \centering
\caption{Clustering performance on graph datasets. 
    The best values are in \textbf{bold}.
    }
    \label{Clustering1}
	\centering
	\begin{tabular*}{\textwidth}{@{\extracolsep{\fill}\quad}lcccccccccc}
    \toprule
       Method &MGAE & DAEGC &DFCN &MVGRL &GDCL &AutoSSL &AGC-DRR &AFGRL &ProGCL &SCGDN  \\\hline  
	\multicolumn{1}{l}{Cora}  \\   
       ~~~~~~ACC\% &43.38±2.11 &70.43±0.36 &36.33±0.49 &70.47±3.70  &70.83±0.47 &63.81±0.57 &40.62±0.55 &26.25±1.24  &57.13±1.23 &\textbf{74.79$\pm$0.38} \\ 
       ~~~~~~NMI\%  &28.78±2.97 &52.89±0.69 &19.36±0.87 &55.57±1.54 &56.30±0.36 &47.62±0.45 &18.74±0.73 &12.36±1.54 &41.02±1.34  &\textbf{56.86$\pm$0.42} \\ 
       ~~~~~~ARI\%  &16.43±1.65 &49.63±0.43 &4.67±2.10 &48.70±3.94  &48.05±0.72 &38.92±0.77 &14.80±1.64 &14.32±1.87 &30.71±2.70  &\textbf{52.61$\pm$0.33} \\ 
       ~~~~~~F1\%    &33.48±3.05 &68.27±0.57 &26.16±0.50 &67.15±1.86  &52.88±0.97 &56.42±0.21 &31.23±0.57 &30.20±1.15  &45.68±1.29 &\textbf{70.42$\pm$0.48} \\  \hline
	\multicolumn{1}{l}{Citeseer}\\ 
	~~~~~~ACC\%  &61.35±0.80 &64.54±1.39 &69.50±0.20 &62.83±1.59  &66.39±0.65 &66.76±0.67 &68.32±1.83 &31.45±0.54  &65.92±0.80  &\textbf{69.62$\pm$0.03} \\  
       ~~~~~~NMI\% &34.63±0.65 &36.41±0.86 &43.90±0.20  &40.69±0.93  &39.52±0.38 &40.67±0.84 &43.28±1.41 &15.17±0.47  &39.59±0.39  &\textbf{44.35$\pm$0.03}  \\ 
       ~~~~~~ARI\%  &33.55±1.18 &37.78±1.24 &45.50±0.30 &34.18±1.73  &41.07±0.96 &38.73±0.55 &45.34±2.33 &14.32±0.78  &36.16±1.11  &\textbf{45.43$\pm$0.04} \\ 
       ~~~~~~F1\%   &57.36±0.82 &62.20±1.32 &64.30±0.20 &59.54±2.17 &61.12±0.70 &58.22±0.68 &64.82±1.60 &30.20±0.71  &57.89±1.98  &\textbf{65.50$\pm$0.06}  \\ \hline
	\multicolumn{1}{l}{AMAP}\\  
	~~~~~~ACC\%  &71.57±2.48 &75.96±0.23 &76.82±0.23 &41.07±3.12 &43.75±0.78 &54.55±0.97 &76.81±1.45 &75.51±0.77  &51.53±0.38 &\textbf{78.91$\pm$0.15}  \\ 
	~~~~~~NMI\% &62.13±2.79 &65.25±0.45 &66.23±1.21 &30.28±3.94  &37.32±0.28 &48.56±0.71 &66.54±1.24 &64.05±0.15  &39.56±0.39 &\textbf{72.53$\pm$0.25}\\
	~~~~~~ARI\%  &48.82±4.57 &58.12±0.24 &58.28±0.74 &18.77±2.34 &21.57±0.51 &26.87±0.34 &60.15±1.56 &54.45±0.48  &34.18±0.89 &\textbf{63.41$\pm$0.21}\\
	~~~~~~F1\%    &68.08±1.76 &69.87±0.54 &71.25±0.31 &32.88±5.50 &38.37±0.29 &54.47±0.83 &71.03±0.64 &69.99±0.34  &31.97±0.44 &\textbf{75.27$\pm$0.18}\\ \hline
	\multicolumn{1}{l}{BAT}\\ 
	~~~~~~ACC\%  &53.59±2.04 &52.67±0.00 &55.73±0.06 &37.56±0.32 &45.42±0.54 &42.43±0.47 &47.79±0.02 &50.92±0.44  &55.73±0.79 &\textbf{74.73$\pm$0.23}    \\ 
	~~~~~~NMI\% &30.59±2.06 &21.43±0.35 &48.77±0.51 &29.33±0.70 &31.70±0.42  &17.84±0.98 &19.91±0.24 &27.55±0.62  &28.69±0.92 &\textbf{52.63$\pm$0.11} \\
	~~~~~~ARI\%  &24.15±1.70 &18.18±0.29 &37.76±0.23 &13.45±0.03 &19.33±0.57 &13.11±0.81 &14.59±0.13 &21.89±0.74  &21.84±1.34 &\textbf{47.65$\pm$0.18} \\
	~~~~~~F1\%   &50.83±3.23 &52.23±0.03 &50.90±0.12 &29.64±0.49 &39.94±0.57 &34.84±0.15 &42.33±0.51 &46.53±0.57  &56.08±0.89 &\textbf{74.49$\pm$0.26}  \\ \hline
	\multicolumn{1}{l}{EAT}\\  
	~~~~~~ACC\% &44.61±2.10 &36.89±0.15 &49.37±0.19 &32.88±0.71 &33.46±0.18 &31.33±0.52 &37.37±0.11 &37.42±1.24  &43.36±0.87 &\textbf{56.52$\pm$0.13}    \\   
	~~~~~~NMI\% &15.60±2.30 &5.57±0.06 &32.90±0.41 &11.72±1.08 &13.22±0.33 &7.63±0.85 &7.00±0.85 &11.44±1.41  &23.93±0.45 &\textbf{32.99$\pm$0.16}  \\ 
	~~~~~~ARI\%  &13.40±1.26 &5.03±0.08 &\textbf{23.25±0.18} &4.68±1.30 &4.31±0.29 &2.13±0.67 &4.88±0.91 &6.57±1.73  &15.03±0.98 &22.89$\pm$0.10  \\
	~~~~~~F1\%   &43.08±3.26 &34.72±0.16 &42.95±0.04 &25.35±0.75 &25.02±0.21 &21.82±0.98 &35.20±0.17 &30.53±1.47  &42.54±0.45 &\textbf{57.63$\pm$0.10}  \\ 
    \bottomrule 
    \end{tabular*}
\end{table*}

\begin{table}[ht]
\setlength{\tabcolsep}{3.3pt}
    \centering
\caption{Clustering performance on CoraFull datasets. 
    The best values are in \textbf{bold}.
    }
    \label{Clustering2}
	\centering
    \begin{tabular*}{\linewidth}{@{\extracolsep{\fill}\,}lcccc}
    \toprule
       Method &ACC\% &NMI\% &ARI\% &F1\%  \\ \hline
      GAE      &29.60±0.81 &45.82±0.75  &17.84±0.86 &25.95±0.75   \\
      %ARGA   &22.07±0.43 &41.28±0.25  &12.38±0.24 &18.85±0.41   \\
      DAEGC &34.35±1.00 &49.16±0.73  &22.60±0.47 &26.96±1.33   \\
      SDCN   &26.67±0.40 &37.38±0.39  &13.63±0.27 &22.14±0.43  \\
      DFCN   &37.51±0.81 &51.30±0.41  &24.46±0.48 &31.22±0.87  \\
      MVGRL&31.52±2.95 &48.99±3.95  &19.11±2.63 &26.51±2.87   \\ 
	DCRN  &38.80±0.60 &51.91±0.35 &25.25±0.49 &31.68±0.76  \\\hline 
      % AGE     &39.62±0.13 &52.38±0.17  &24.46±0.23 &27.95±0.19   \\
	SCGDN &\textbf{40.13$\pm$0.4}1 &\textbf{54.15$\pm$0.08} &\textbf{26.97$\pm$0.87} &\textbf{34.77$\pm$0.26}  \\ 
    \bottomrule 
    \end{tabular*}
\end{table}

\textbf{Experimental setting.\,}
Our proposed SCGDN is trained using Adam~\cite{kingma2014adam}, in which the learning rate is set to 1e-3 for AttM, and 1e-5 for DiFM, respectively.
We first train the model in an unsupervised manner, then perform evaluations on the learned representations.
For all experiments, we report the mean accuracy with a standard deviation through 10 random initializations.
About building the t-distribution $k$NN graph, we fix the standard deviation $\sigma=0.5$ and the degrees of freedom $v=1$, and set the dimension of hidden layer in each dataset to 512, except for BAT dataset. Because the feature dimension of BAT dataset is 81, less than 512.
Detailed hyperparameter setting is in Appendix~\ref{Hyperparameters}.

We implement SCGDN with PyTorch and all experiments are conducted on NVIDIA RTX 3090 GPUs. 
We use widely used four metrics to evaluate clustering performance, including Accuracy (ACC), Normalized Mutual Information (NMI), Adjusted Rand Index (ARI), and F1-score (F1).

\subsection{Node Clustering}
In this section, we compare SCGDN with three types of deep clustering methods and report the node clustering results in Tables~\ref{Clustering1} and \ref{Clustering2}.

Compared to adversarial and generative methods, it can be viewed that SCGDN outperforms the baselines. This is because SCGDN which inherits the advantage of the contrastive methods has more available supervision information. Whether in large or small datasets, we have empirically verified the superior performance of SCGDN compared with other contrastive methods. This is because SCGDN captures intrinsic category information due to high-quality sampling. On AMAP and CoraFull datasets which are the larger benchmark, SCGDN also achieves the best performance. It is worth mentioned that we empirically find that representation of 64 dimensions is better than baselines on BAT dataset.

\subsection{Ablation studies}\label{Abl}
In this section, we first conduct ablation studies to verify the effectiveness of each component in our proposed loss function on five benchmark datasets as shown in Table~\ref{Ablation1}.
In addition, we conduct another ablation studies to verify the effectiveness of the proposed AttM and DiFM on Cora and Citeseer datasets as shown in Table~\ref{Ablation2}.
Last, we conduct ablation studies to verify the effectiveness of the negative sampling strategy.

\textbf{Effectiveness of $\mathcal{L}_{p}$, $\mathcal{L}_{n}$ and $\mathcal{L}_{r}$.} 
To see the impact of the positive and negative contrastive term and the redundancy reduction term, we conduct the ablation studies which have different types of objective functions. 
The first variant, " $\mathcal{L}_{r}$", only uses the redundancy reduction term. 
The second variant, which we refer to as "(w/o)$\mathcal{L}_{n}$", uses the positive contrastive term and the redundancy reduction term to guide the network parameter updates. 
The "(w/o)$\mathcal{L}_{r}$"  uses the positive contrastive term and the negative contrastive term. 
From these results, we have three findings as follows.
(1) The positive and negative contrastive terms provide more supervision information implicitly.
(2) The redundancy reduction term helps reduce feature redundancy in potential spaces to obtain a more discriminative representation. 
(3) The negative contrastive term further boosts the performance of clustering by pushing the distance of the negative samples. 
Overall, these experiments are sufficient to illustrate the necessity of each component in our proposed loss function.

\begin{table}[ht]
\setlength{\tabcolsep}{3.3pt}
    \centering
\caption{Ablation studies of functions on graph datasets. 
    The best values are in \textbf{bold}.
    }
    \label{Ablation1}
	\centering
	\begin{tabular*}{\linewidth}{@{\extracolsep{\fill}\quad}lcccc}
    \toprule
       Dataset &$\mathcal{L}_{r}$ &(w/o)$\mathcal{L}_{n}$ & (w/o)$\mathcal{L}_{r}$  &SCGDN  \\\hline  
	\multicolumn{1}{l}{Cora}  \\   
       ~~~~~~ACC\%  &47.47$\pm$2.18  &74.04$\pm$0.09  &74.23$\pm$0.04  &\textbf{74.79$\pm$0.38} \\ 
       ~~~~~~NMI\%  &31.73$\pm$2.09  &55.75$\pm$0.13  &56.30$\pm$0.12  &\textbf{56.86$\pm$0.42} \\ 
       ~~~~~~ARI\%   &14.28$\pm$2.28  &51.56 $\pm$ 0.16  &52.08$\pm$0.13  &\textbf{52.61$\pm$0.33} \\ 
       ~~~~~~F1\%     &43.23$\pm$3.53  &69.63 $\pm$ 0.07 &69.75$\pm$0.06   &\textbf{70.42$\pm$0.48} \\  \hline
	\multicolumn{1}{l}{Citeseer} \\
	~~~~~~ACC\%  &64.14$\pm$0.08  &69.47$\pm$0.03  &69.55$\pm$0.02  &\textbf{69.62$\pm$0.03} \\ 
	~~~~~~NMI\%  &39.51$\pm$0.05  &44.03$\pm$0.08  &44.09$\pm$0.01 &\textbf{44.35$\pm$0.03}  \\ 
	~~~~~~ARI\%   &37.49$\pm$0.09 &45.15$\pm$0.06  &45.30$\pm$0.03 &\textbf{45.43$\pm$0.04} \\ 
	~~~~~~F1\%     &60.50$\pm$0.03  &65.01$\pm$0.03  &65.49$\pm$0.04 &\textbf{65.50$\pm$0.06}  \\ \hline
	\multicolumn{1}{l}{AMAP}\\
	~~~~~~ACC\% &61.57$\pm$1.36  &78.85$\pm$0.07  &78.03$\pm$0.01 &\textbf{78.91$\pm$0.15}  \\ 
	~~~~~~NMI\% &48.79$\pm$1.69  &72.34$\pm$0.19  &70.96$\pm$0.02  &\textbf{72.53$\pm$0.25}\\  
	~~~~~~ARI\%  &33.94$\pm$2.35  &63.28$\pm$0.11  &62.26$\pm$0.06  &\textbf{63.41$\pm$0.21}\\
	~~~~~~F1\%    &50.35$\pm$3.50  &75.21$\pm$0.26  &74.15$\pm$0.03  &\textbf{75.27$\pm$0.18}\\ \hline
	\multicolumn{1}{l}{BAT}\\
	~~~~~~ACC\%  &54.50$\pm$0.37  &65.88$\pm$5.91  &74.27$\pm$0.35 &\textbf{74.73$\pm$0.23}    \\ 
	~~~~~~NMI\%  &42.77$\pm$1.10  &46.49$\pm$3.47  &52.40$\pm$0.17  &\textbf{52.63$\pm$0.11} \\
	~~~~~~ARI\%  &29.49$\pm$0.56  &38.71$\pm$5.21  &47.28$\pm$0.28  &\textbf{47.65$\pm$0.18} \\
	~~~~~~F1\%    &47.19$\pm$0.31  &64.65$\pm$6.81   &73.96$\pm$0.40  &\textbf{74.49$\pm$0.26}  \\ \hline
	\multicolumn{1}{l}{EAT}\\
	~~~~~~ACC\% &52.23$\pm$0.23  &54.34$\pm$0.31 &54.64$\pm$0.19  &\textbf{56.52$\pm$0.13}    \\ 
	~~~~~~NMI\% &33.02$\pm$0.10   &34.66$\pm$0.19 &\textbf{34.71$\pm$0.32} &32.99$\pm$0.16  \\ 
	~~~~~~ARI\%  &22.96$\pm$0.17  &23.58$\pm$0.21 &\textbf{23.73$\pm$0.19}  &22.89$\pm$0.10  \\
	~~~~~~F1\%    &53.05$\pm$0.21  &54.96$\pm$0.28  &55.20$\pm$0.18  &\textbf{57.63$\pm$0.10}  \\
    \bottomrule 
    \end{tabular*}
\end{table}

\textbf{Effectiveness of AttM and DiFM.} 
Here, we denote "Ours${_{GCN}}$", "Ours${_{AttM+GCN}}$", and "Ours", as the model of GCN, the model of AttM and GCN, and SCGDN, respectively.
From the results of Table~\ref{Ablation2} and Table~\ref{Ablation3}, we have three observations as follows.
(1) Using our proposed graph contrastive paradigm, good results can also be achieved under the framework of GCN.
(2) Our proposed AttM better aggregates structure and feature information.
(3) With an understanding of the nature of graph problems, our proposed DiFM does provide a better diffusion of learned representations.
To sum up, these experiments validate the effectiveness of the model framework.

\begin{table}[ht]
\setlength{\tabcolsep}{3.3pt}
    \centering
\caption{Ablation studies of models on Cora and Citeseer datasets. 
    The best values are in \textbf{bold}.
    }
    \label{Ablation2}
	\centering
    \begin{tabular*}{\linewidth}{@{\extracolsep{\fill}\,}l|ccc|ccc}
    \toprule
       Method  &\multicolumn{3}{c|}{Cora} &\multicolumn{3}{c}{Citeseer}\\
      ~    &ACC\% &NMI\%  &F1\%  &ACC\% &NMI\%  &F1\% \\ \hline
      Ours${_{GCN}}$   &67.47 &49.58 &63.65   &64.50 &37.12 &59.30  \\  
      Ours${_{AttM+GCN}}$   &73.01 &53.85 &63.91   &69.37 &44.26 &60.68  \\ 
	 Ours &\textbf{74.79} &\textbf{56.86} &\textbf{70.42} &\textbf{69.62} &\textbf{44.35} &\textbf{65.50} \\ 
    \bottomrule 
    \end{tabular*}
\end{table}

\textbf{Effectiveness of the negative sampling strategy.} 
In these ablation studies, "DGI" denotes the model of GCN with the classic contrastive method. And "Ours$_{random}$" denotes randomly generating negative samples in SCGDN.
From the results of Table~\ref{Ablation3}, we have the observation as follows: the negative sampling strategy could improve the performance of SCGDN, and its performance exceeds that of DGI and "Ours$_{random}$".
Overall, these experiments are sufficient to illustrate the effectiveness of the negative sampling strategy.

\begin{table}[ht]
\setlength{\tabcolsep}{3.3pt}
    \centering
\caption{Ablation studies of models on Cora and Citeseer datasets. 
    The best values are in \textbf{bold}.
    }
    \label{Ablation3}
	\centering
    \begin{tabular*}{\linewidth}{@{\extracolsep{\fill}\,}l|ccc|ccc}
    \toprule
       Method &\multicolumn{3}{c|}{Cora} &\multicolumn{3}{c}{Citeseer}\\
      ~    &ACC\% &NMI\%  &F1\%  &ACC\% &NMI\%  &F1\% \\ \hline
      DGI   &71.81 &54.09  &69.88 &68.60 &43.75 &64.64   \\
	Ours${_{random}}$  &73.45 &\textbf{56.96} &69.76   &69.00 &43.40 &64.18  \\
	 Ours &\textbf{74.79} &56.86 &\textbf{70.42} &\textbf{69.62} &\textbf{44.35} &\textbf{65.50} \\ 
    \bottomrule 
    \end{tabular*}
\end{table}

\begin{figure}[!ht]
\centering
	  \includegraphics[width=0.15\textwidth]{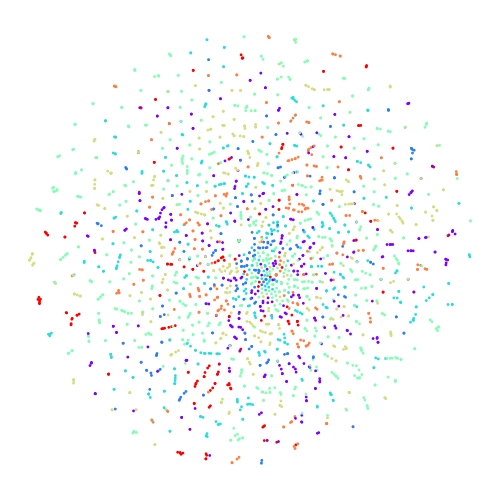}
	  \includegraphics[width=0.15\textwidth]{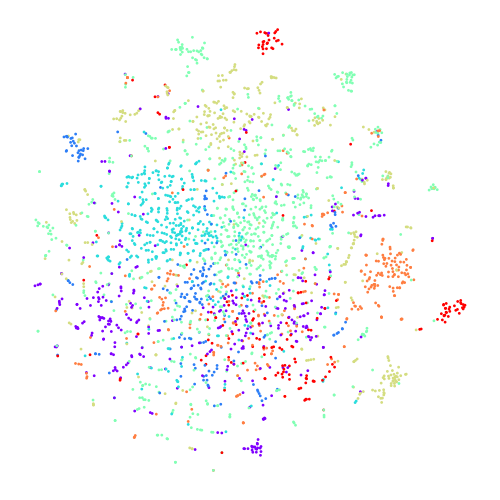}
	  \includegraphics[width=0.15\textwidth]{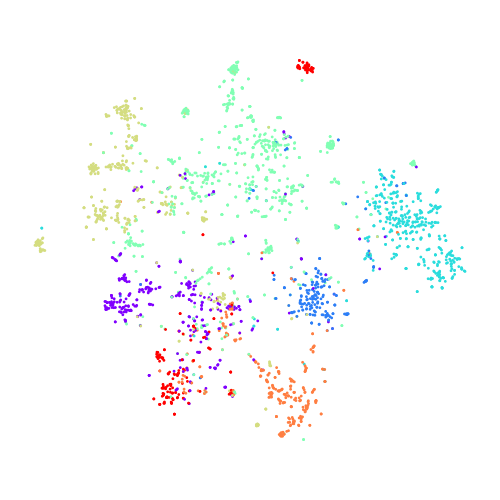}\\
	\subcaptionbox*{$\mathbf{W}$} {
	 \includegraphics[width=0.15\textwidth]{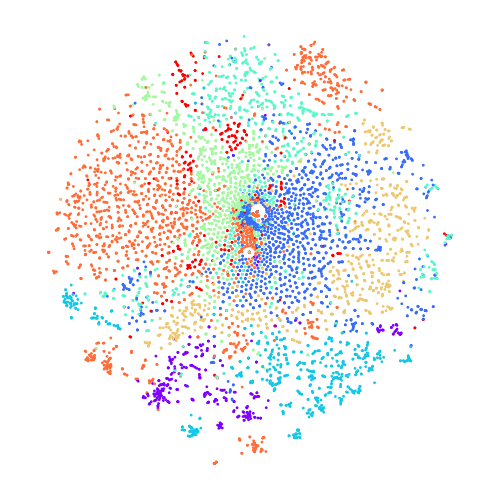}}
	%\subcaptionbox*{$X$} {
	% \includegraphics[width=0.11\textwidth]{figures/amap_feature.png}}
	\subcaptionbox*{$\mathbf{W}^{knn}$} {
	 \includegraphics[width=0.15\textwidth]{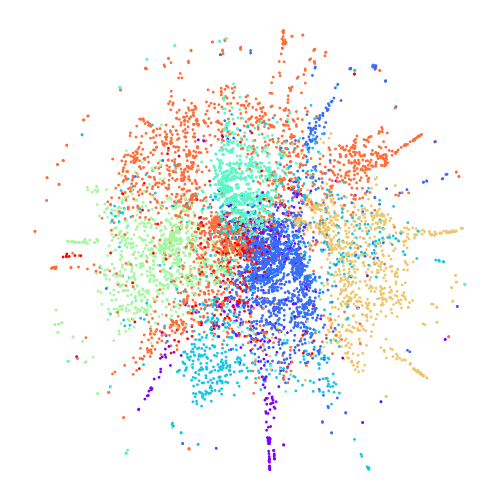}}
	\subcaptionbox*{Embeeding} {
	  \includegraphics[width=0.15\textwidth]{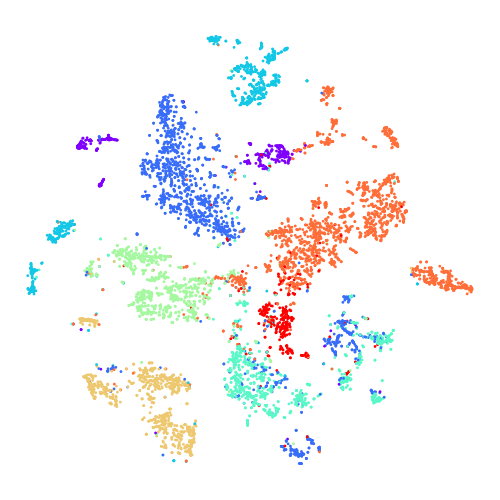}}
        \caption{2D t-SNE visualization on two benchmark datasets. The first row and second row correspond to Cora and AMAP datasets, respectively.}
  \label{Vis}
\end{figure}

\begin{figure*}[ht]
\centering
	  \includegraphics[width=0.245\textwidth]{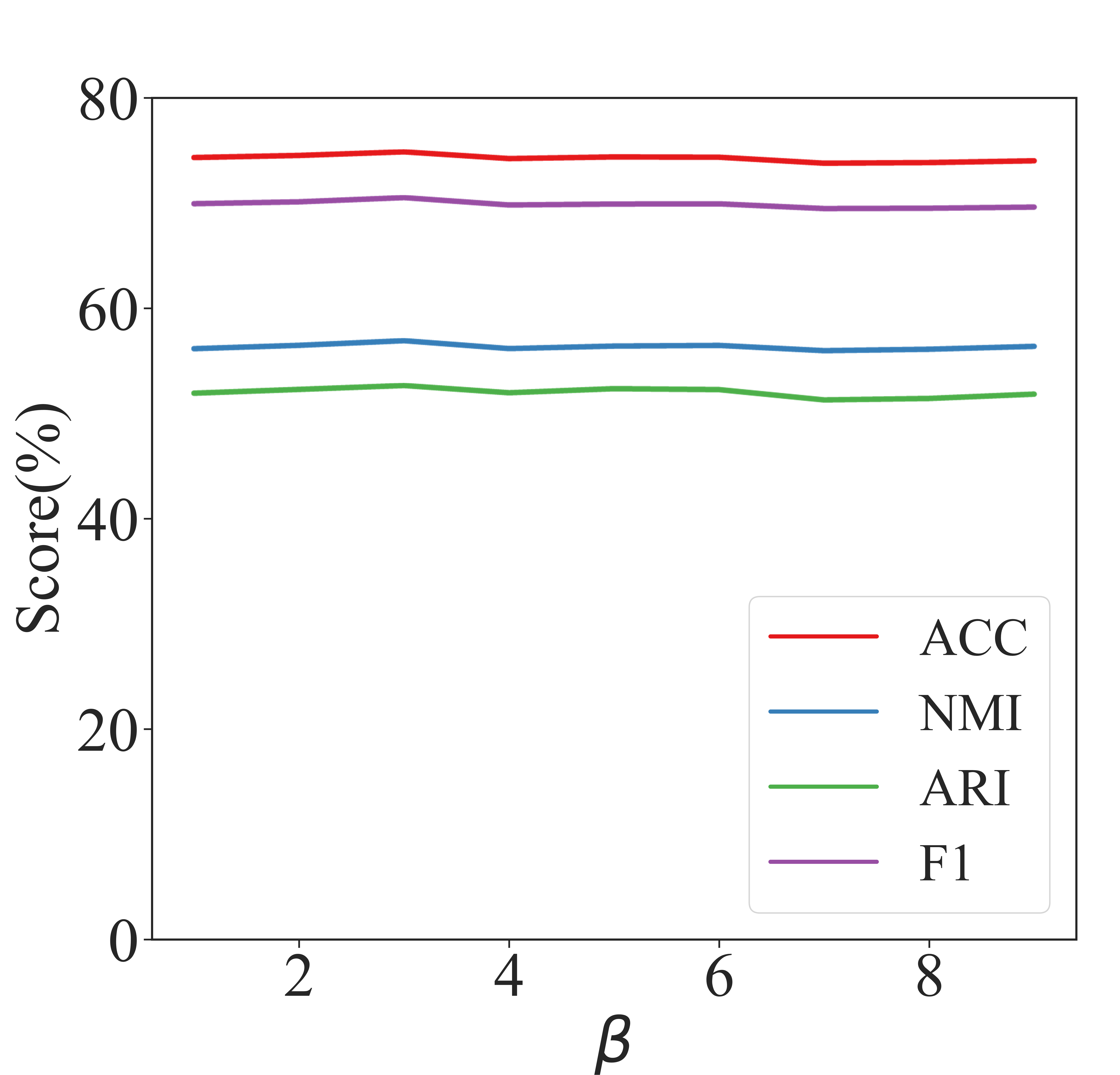}
	  \includegraphics[width=0.245\textwidth]{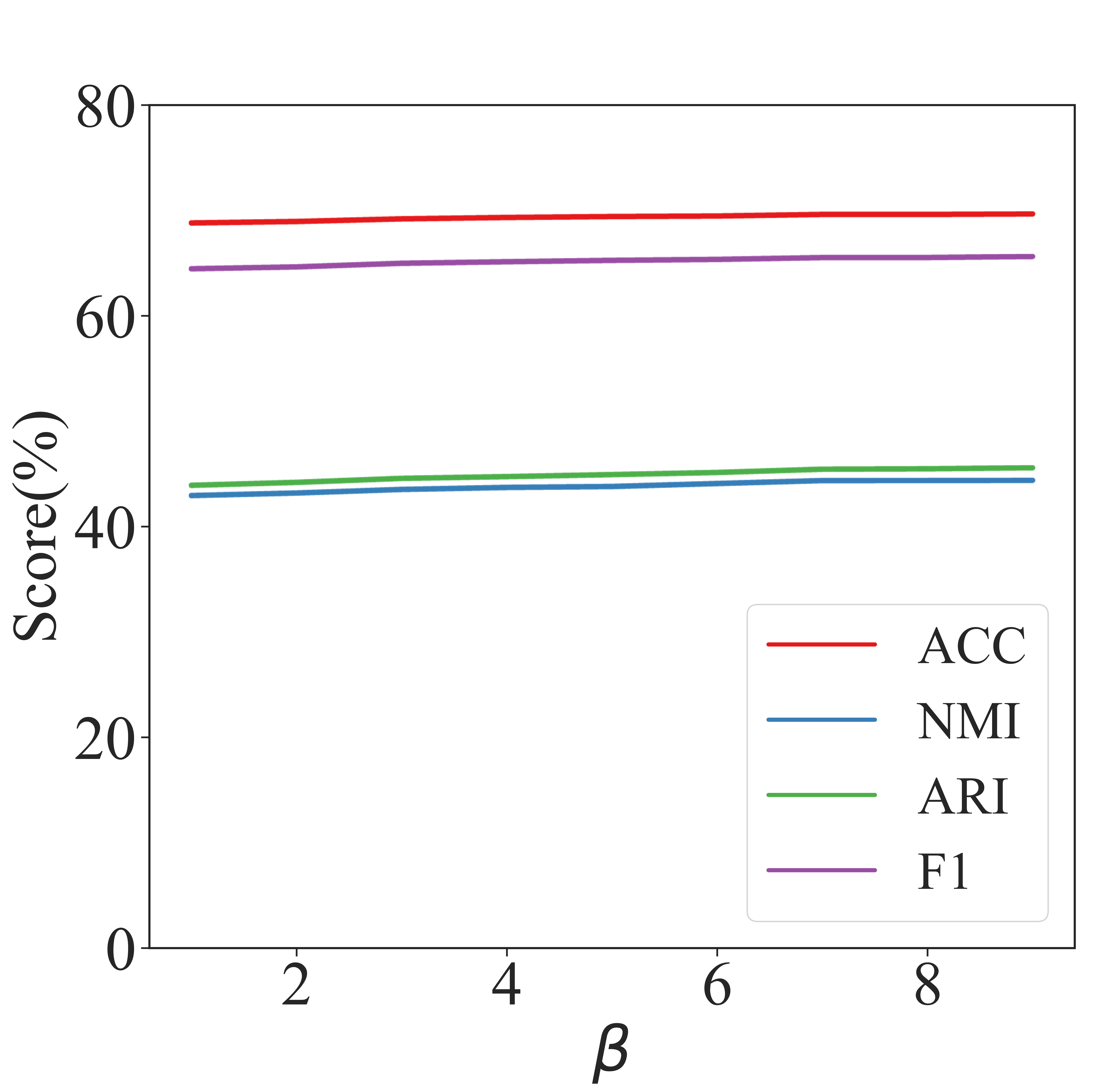}
	  \includegraphics[width=0.245\textwidth]{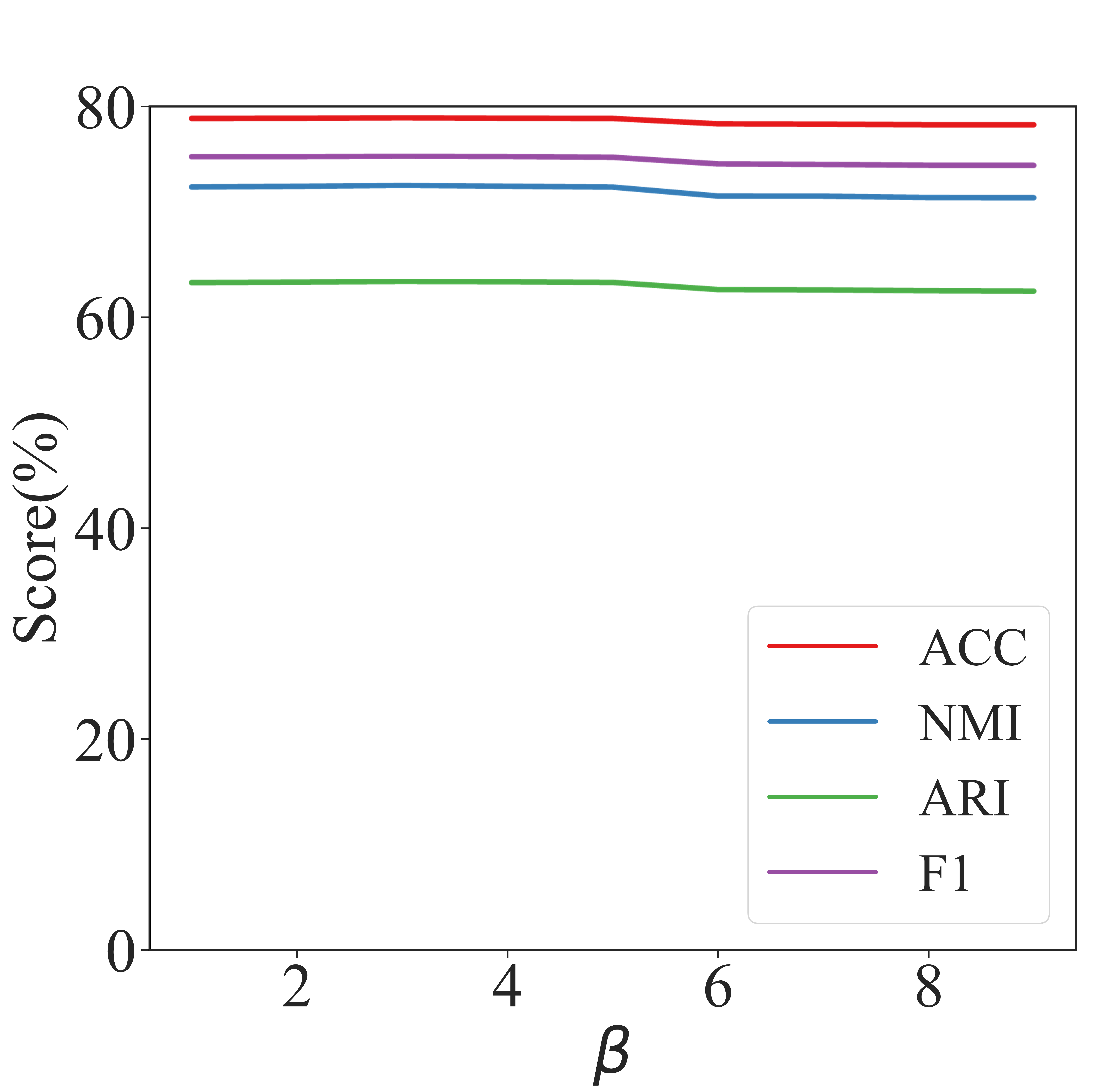}
	  \includegraphics[width=0.245\textwidth]{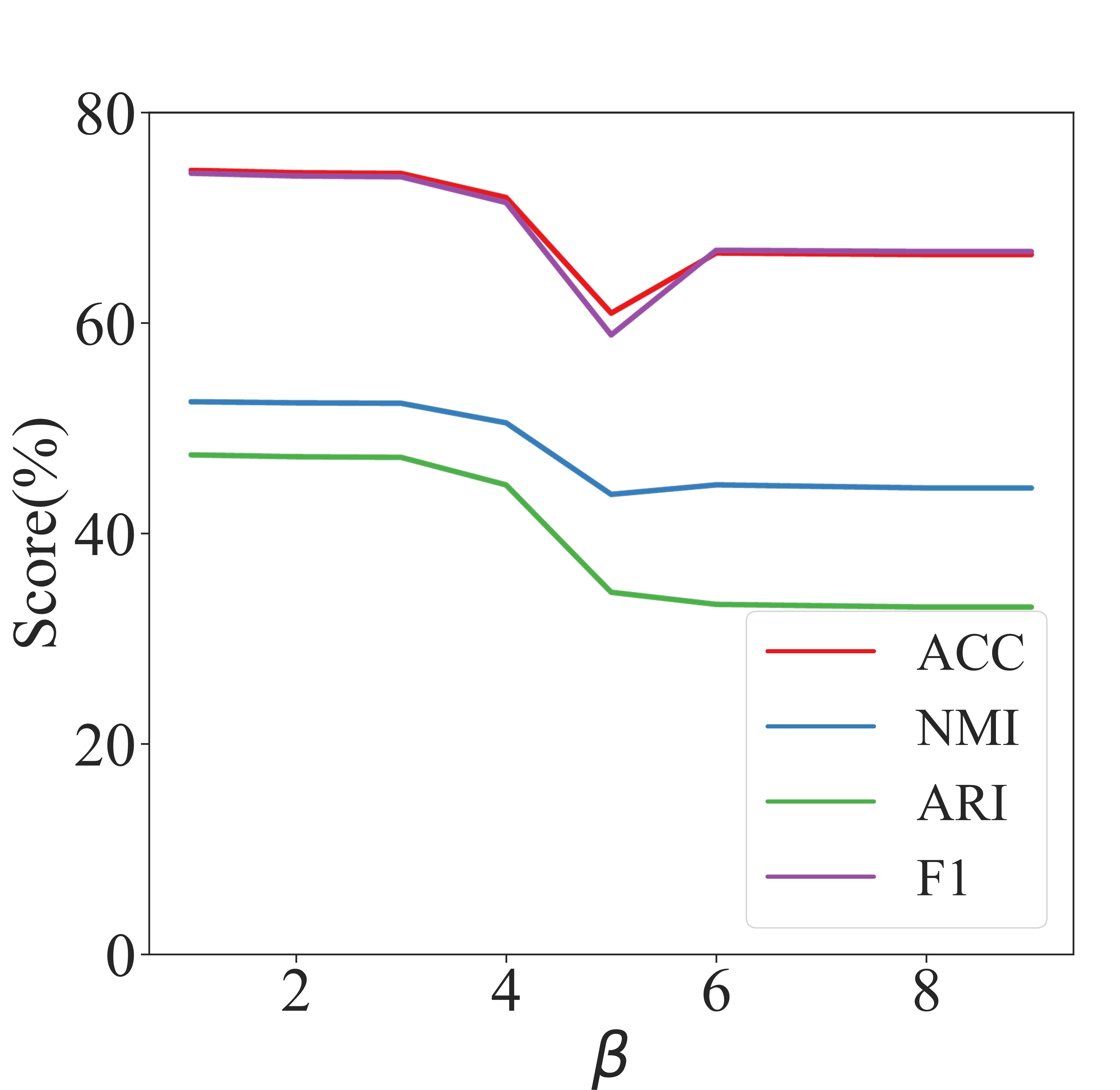}\\
	\subcaptionbox*{Cora} {
	  \includegraphics[width=0.24\textwidth]{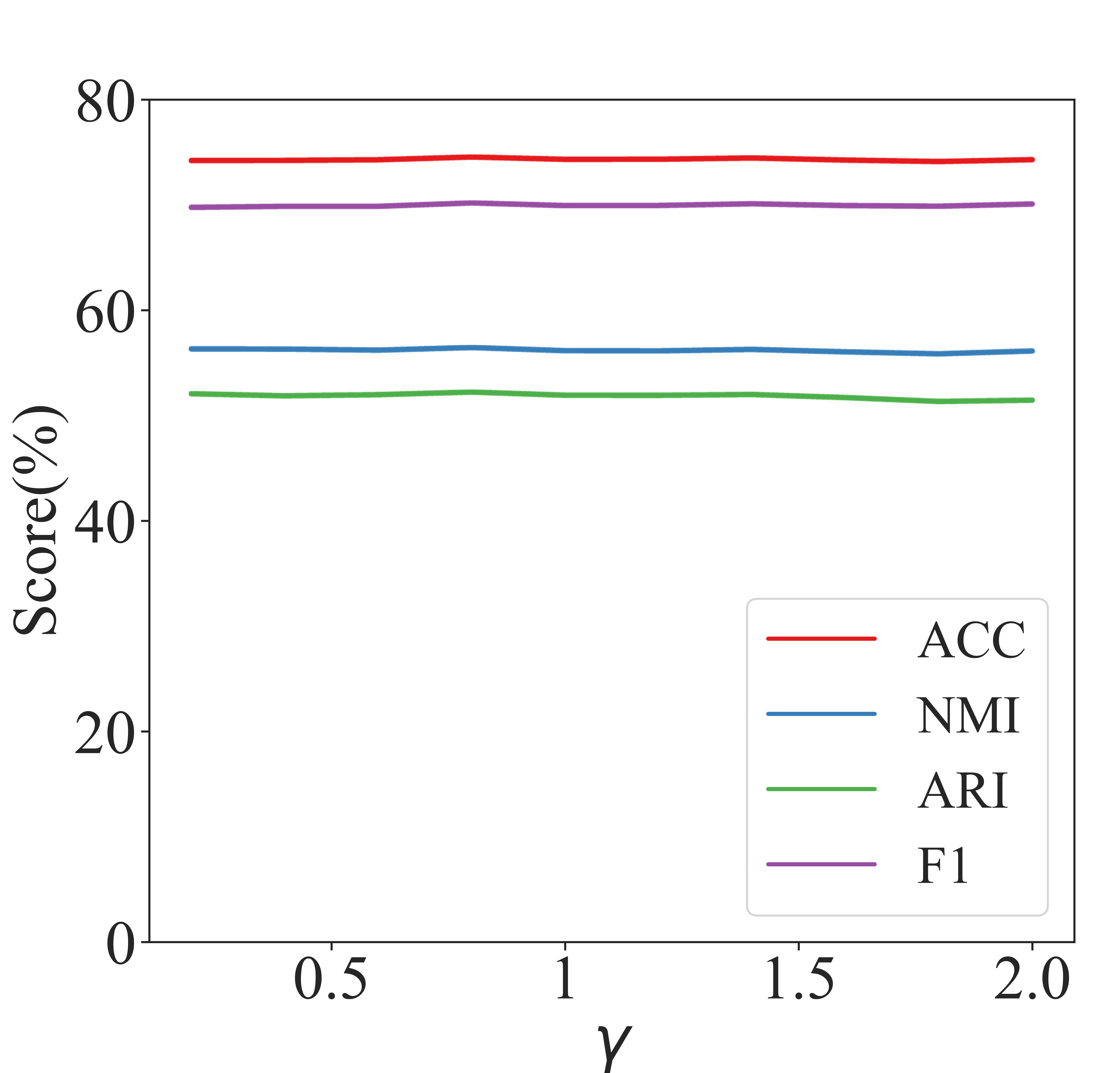}}
	\subcaptionbox*{Citeseer} {
	  \includegraphics[width=0.24\textwidth]{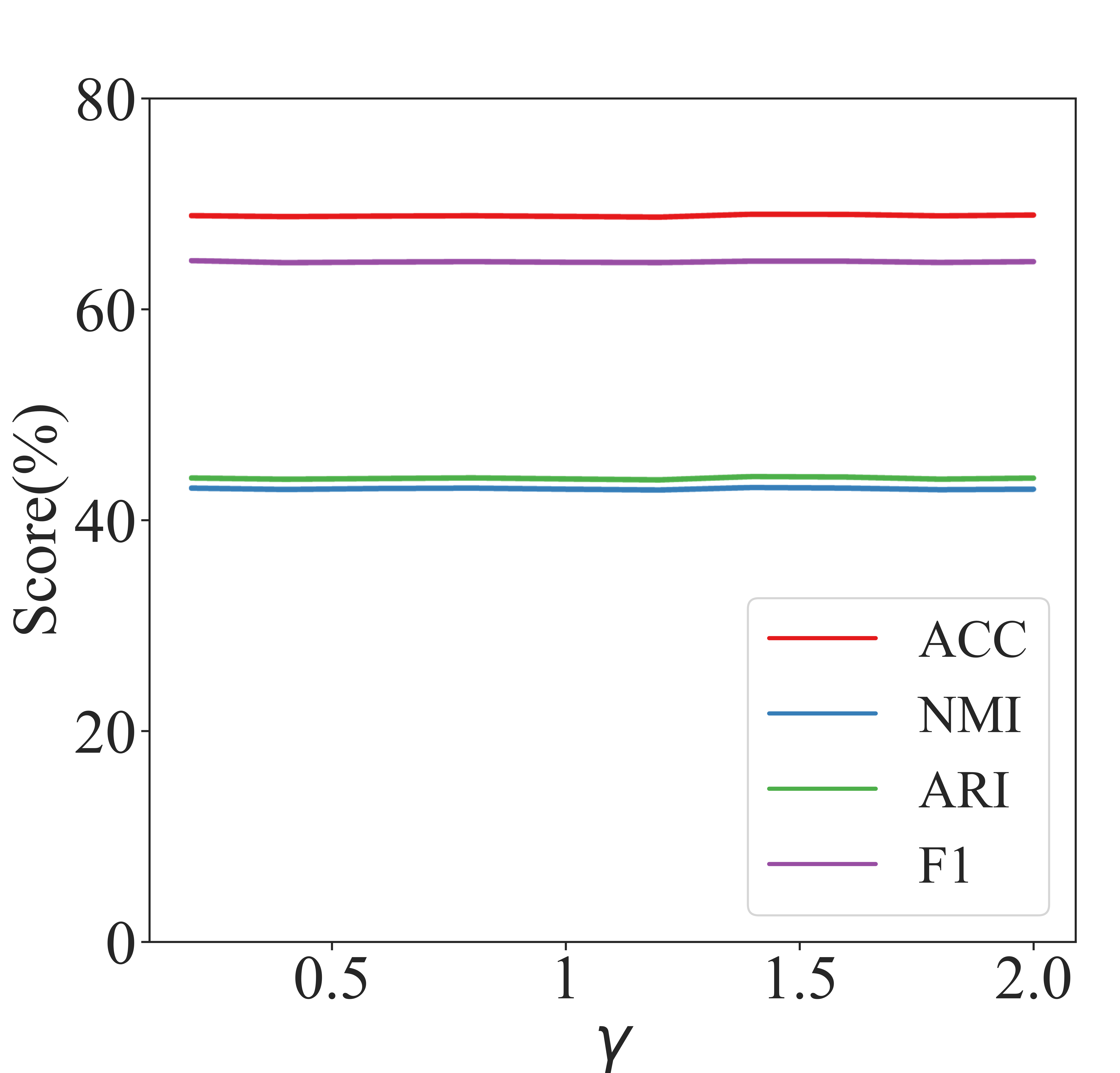}}
	\subcaptionbox*{AMAP} {
	  \includegraphics[width=0.24\textwidth]{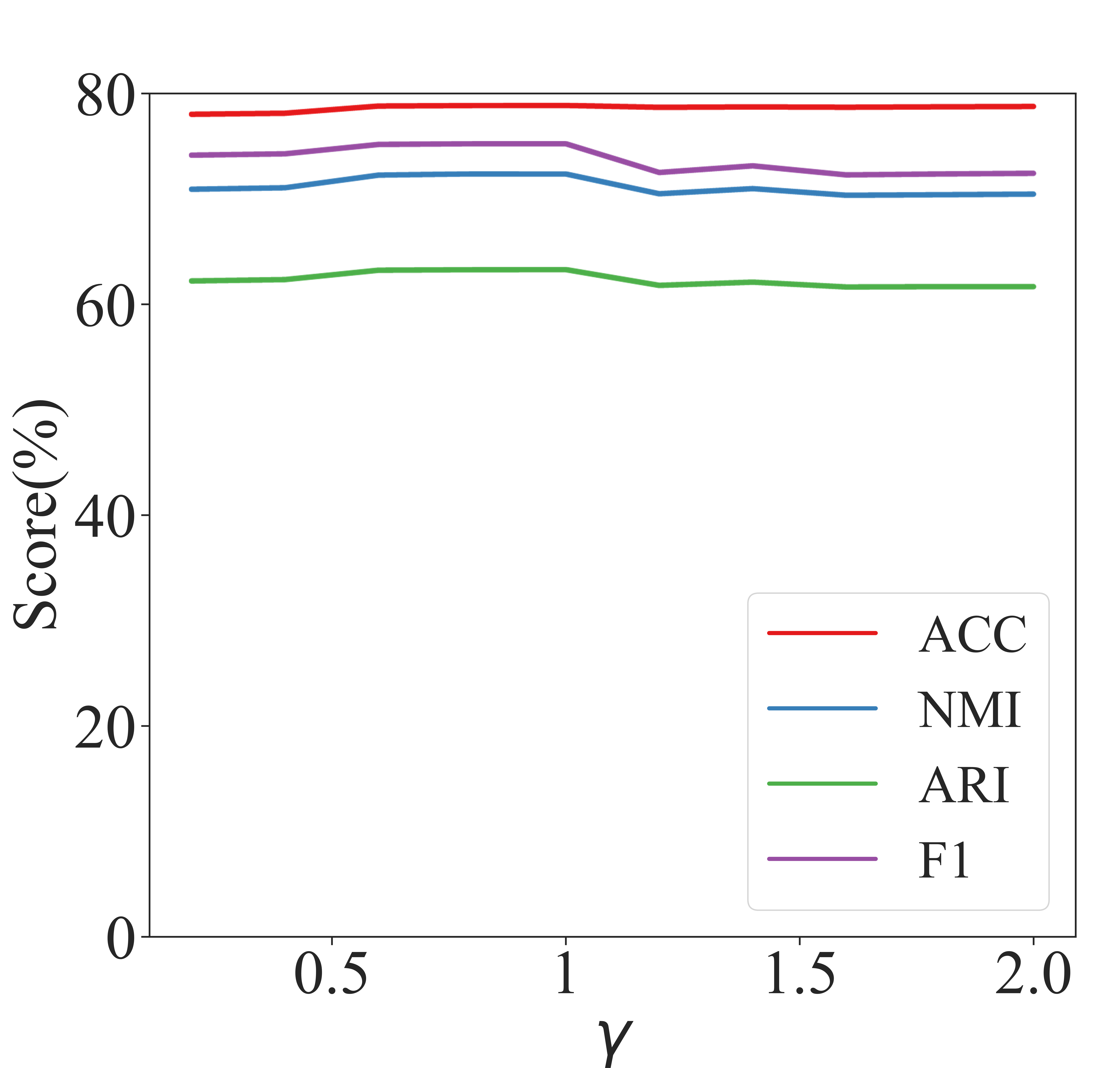}}
	\subcaptionbox*{BAT} {
	  \includegraphics[width=0.24\textwidth]{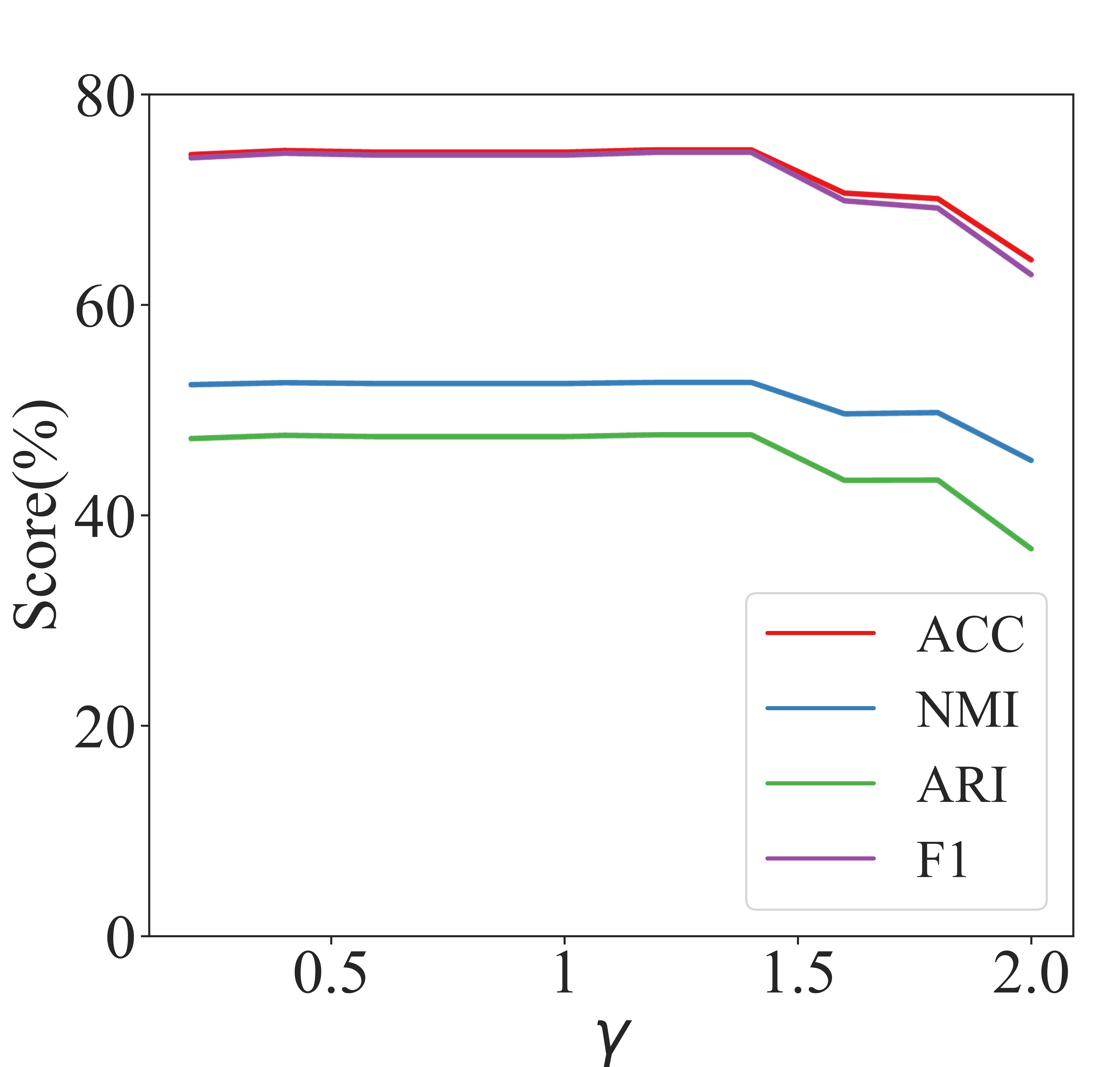}}
        \caption{Parameter sensitivity of $\beta$ and $\gamma$ on four benchmark datasets. The first row and second row correspond to sensitivity of $\beta$ and  $\gamma$, respectively.}
  \label{sensitivity}
\end{figure*} 

\subsection{Analysis}
\subsubsection{Visualization analysis}
In this part, we visualize the distribution of the learned representations to provide a more intuitive understanding of the learned node representations of SCGDN on Cora and AMAP datasets via the t-SNE algorithm~\cite{van2008visualizing}. The t-SNE focuses on data points that are relatively close together in high-dimensional space. Therefore, the embedding local structure can be observed more clearly and intuitively. The color represents the node label, and each point represents a node.
From Fig~\ref{Vis}, we have the following observations:
(1) The raw data of Cora dataset lacks obvious clustering, and it appears chaotic. However, the learned node representations by SCGDN show more tightly grouped nodes. This signifies that SCGDN captures more fine-grained class information. 
(2) Although there is the class information in the raw data of the AMAP dataset, different classes are not clearly separated. From the learned node representations of SCGDN, we observe that there is a significant gap between different categories, and the same category is more closely spaced. This indicates that the optimization objective has narrowed the distance between the same class and widened the distance between different classes.

\subsubsection{Convergence analysis}
\begin{figure}[!ht]
\centering
	  \includegraphics[width=0.236\textwidth]{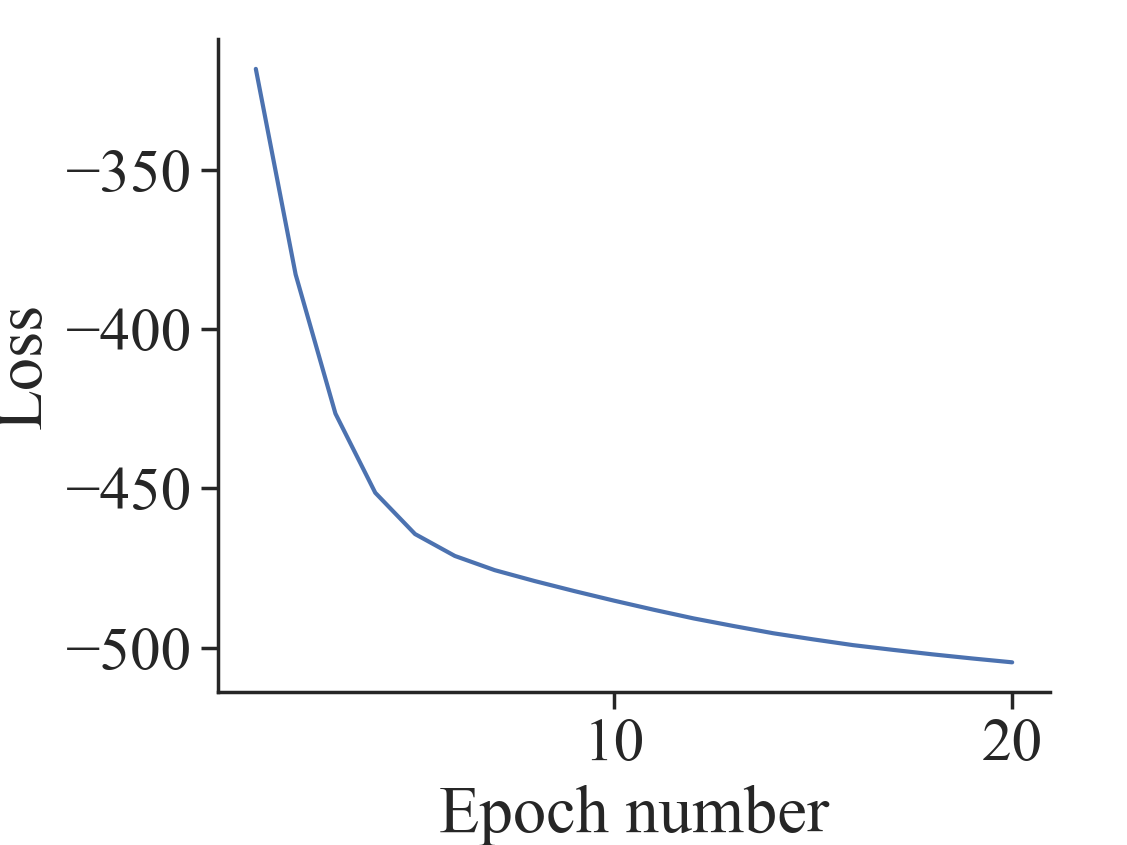}
	  \includegraphics[width=0.236\textwidth]{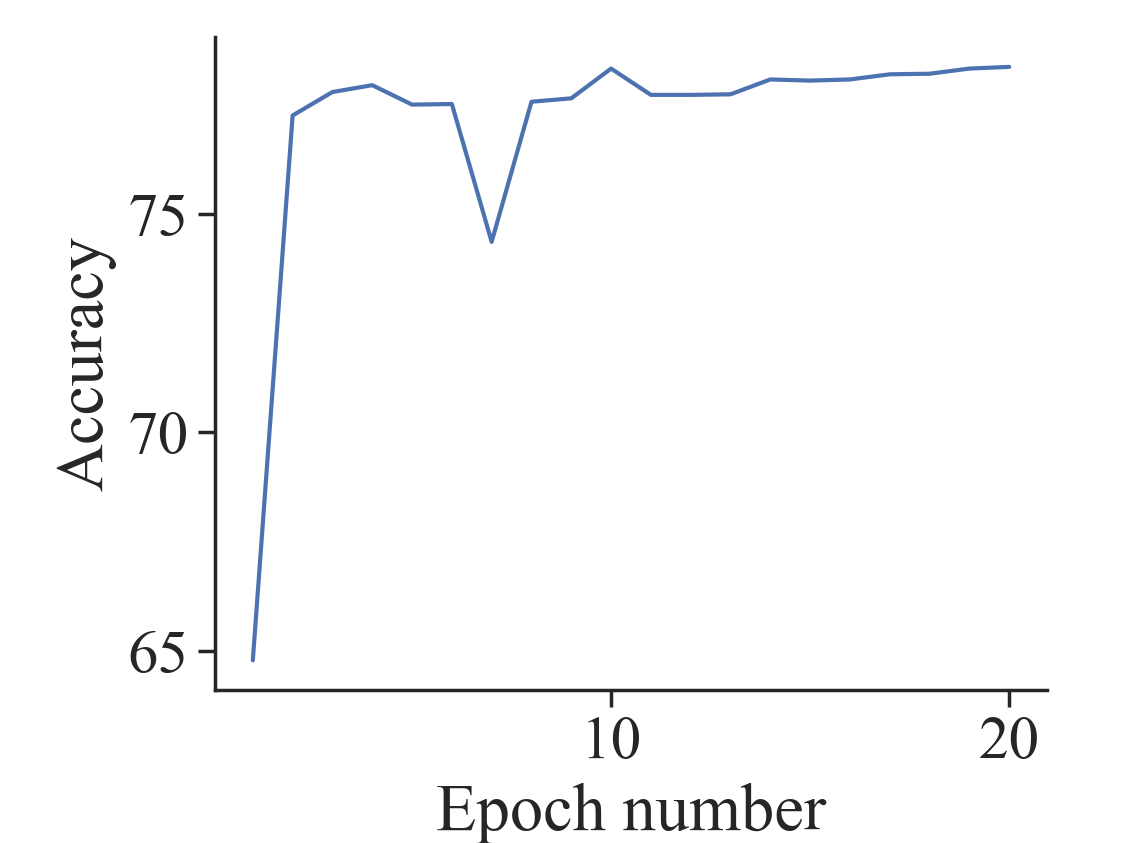}
        \caption{Convergence analysis of the proposed loss on AMAP dataset.}
  \label{Convergence}
\end{figure}

To further evaluate the performance of the proposed loss function, we conducted experiments to analyze the convergence of the loss. Specifically, we plotted the trend of the loss and the ACC metric on AMAP dataset, as shown in Fig.~\ref{Convergence}. 
As the loss decreases, the accuracy rate gradually increases and tends to be stable. 

\subsubsection{Parameter sensitivity analysis}
For unsupervised contrastive learning methods, parameter insensitivity is vital for enhancing their stability. For $\beta$, we set it in \{0.5, 1, 2, 3, 4, 5, 6, 7, 8, 9\}, and for $\gamma$, we set it in \{0.2, 0.4, 0.6, 0.8, 1, 1.2, 1.4, 1.6 1.8, 2\}.
From Fig.~\ref{sensitivity}, conclusion is deduced that SCGDN is not sensitive to the factors $\beta$ and $\gamma$, which enhances the practicality in real-world applications.

\section{Conclusion}
In this work, we propose a novel Self-Contrastive Graph Diffusion Network (SCGDN), an augmentation-free and free pre-training method. SCGDN effectively balances between preserving high-order structure information and avoiding overfitting. 
First, we design an efficient and effective model framework, including two main parts, \textit{\ie}, Attentional Module (AttM) and Diffusion Module (DiFM). 
Specifically, AttM aggregates higher-order structure and feature information to get a excellent embedding. DiFM balances the stationary state of each node in the graph through diffusion learning. Meanwhile, we introduced a novel graph contrastive learning paradigm that conducts contrastive learning with the proposed high-quality negative sampling strategy and without multiview. Compared with other contrastive methods, SCGDN not only further improves the discriminative capability of the learned representations, but also utilizes intrinsic feature information and higher-order structure. In light of extensive experiments on six benchmark datasets, the results indicated the effectiveness and superiority of the 
proposed SCGDN. In the future, we plan to apply the proposed method to solve more real applications.

\bibliographystyle{ACM-Reference-Format}
\bibliography{ID533}


\begin{thebibliography}{40}

%%% the Web address.

\ifx \showCODEN    \undefined \def \showCODEN     #1{\unskip}     \fi
\ifx \showDOI      \undefined \def \showDOI       #1{#1}\fi
\ifx \showISBNx    \undefined \def \showISBNx     #1{\unskip}     \fi
\ifx \showISBNxiii \undefined \def \showISBNxiii  #1{\unskip}     \fi
\ifx \showISSN     \undefined \def \showISSN      #1{\unskip}     \fi
\ifx \showLCCN     \undefined \def \showLCCN      #1{\unskip}     \fi
\ifx \shownote     \undefined \def \shownote      #1{#1}          \fi
\ifx \showarticletitle \undefined \def \showarticletitle #1{#1}   \fi
\ifx \showURL      \undefined \def \showURL       {\relax}        \fi
% invisible to TeX
\providecommand\bibfield[2]{#2}
\providecommand\bibinfo[2]{#2}
\providecommand\natexlab[1]{#1}
\providecommand\showeprint[2][]{arXiv:#2}

\bibitem[Belkin and Niyogi(2003)]%
        {belkin2003laplacian}
\bibfield{author}{\bibinfo{person}{Mikhail Belkin} {and}
  \bibinfo{person}{Partha Niyogi}.} \bibinfo{year}{2003}\natexlab{}.
\newblock \showarticletitle{Laplacian eigenmaps for dimensionality reduction
  and data representation}.
\newblock \bibinfo{journal}{\emph{Neural computation}} \bibinfo{volume}{15},
  \bibinfo{number}{6} (\bibinfo{year}{2003}), \bibinfo{pages}{1373--1396}.
\newblock


\bibitem[Bo et~al\mbox{.}(2020)]%
        {bo2020structural}
\bibfield{author}{\bibinfo{person}{Deyu Bo}, \bibinfo{person}{Xiao Wang},
  \bibinfo{person}{Chuan Shi}, \bibinfo{person}{Meiqi Zhu},
  \bibinfo{person}{Emiao Lu}, {and} \bibinfo{person}{Peng Cui}.}
  \bibinfo{year}{2020}\natexlab{}.
\newblock \showarticletitle{Structural Deep Clustering Network}. In
  \bibinfo{booktitle}{\emph{WWW}}. \bibinfo{pages}{1400--1410}.
\newblock


\bibitem[Boudiaf et~al\mbox{.}(2020)]%
        {boudiaf2020unifying}
\bibfield{author}{\bibinfo{person}{Malik Boudiaf},
  \bibinfo{person}{J{\'e}r{\^o}me Rony}, \bibinfo{person}{Imtiaz~Masud Ziko},
  \bibinfo{person}{Eric Granger}, \bibinfo{person}{Marco Pedersoli},
  \bibinfo{person}{Pablo Piantanida}, {and} \bibinfo{person}{Ismail {Ben
  Ayed}}.} \bibinfo{year}{2020}\natexlab{}.
\newblock \showarticletitle{A unifying mutual information view of metric
  learning: cross-entropy vs. pairwise losses}. In
  \bibinfo{booktitle}{\emph{ECCV}}. \bibinfo{pages}{548--564}.
\newblock


\bibitem[Chamberlain et~al\mbox{.}(2021a)]%
        {chamberlain2021blend}
\bibfield{author}{\bibinfo{person}{Benjamin~Paul Chamberlain},
  \bibinfo{person}{James Rowbottom}, \bibinfo{person}{Davide Eynard},
  \bibinfo{person}{Francesco Di~Giovanni}, \bibinfo{person}{Dong Xiaowen},
  {and} \bibinfo{person}{Michael~M Bronstein}.}
  \bibinfo{year}{2021}\natexlab{a}.
\newblock \showarticletitle{Beltrami Flow and Neural Diffusion on Graphs}. In
  \bibinfo{booktitle}{\emph{NeurIPS}}, Vol.~\bibinfo{volume}{34}.
  \bibinfo{pages}{1594--1609}.
\newblock


\bibitem[Chamberlain et~al\mbox{.}(2021b)]%
        {chamberlain2021grand}
\bibfield{author}{\bibinfo{person}{Benjamin~Paul Chamberlain},
  \bibinfo{person}{James Rowbottom}, \bibinfo{person}{Maria Goronova},
  \bibinfo{person}{Stefan Webb}, \bibinfo{person}{Emanuele Rossi}, {and}
  \bibinfo{person}{Michael~M Bronstein}.} \bibinfo{year}{2021}\natexlab{b}.
\newblock \showarticletitle{GRAND: Graph Neural Diffusion}. In
  \bibinfo{booktitle}{\emph{ICML}}, Vol.~\bibinfo{volume}{139}.
  \bibinfo{pages}{1407--1418}.
\newblock


\bibitem[Chen et~al\mbox{.}(2022)]%
        {chen2022optimization}
\bibfield{author}{\bibinfo{person}{Qi Chen}, \bibinfo{person}{Yifei Wang},
  \bibinfo{person}{Yisen Wang}, \bibinfo{person}{Jiansheng Yang}, {and}
  \bibinfo{person}{Zhouchen Lin}.} \bibinfo{year}{2022}\natexlab{}.
\newblock \showarticletitle{Optimization-Induced Graph Implicit Nonlinear
  Diffusion}. In \bibinfo{booktitle}{\emph{ICML}}, Vol.~\bibinfo{volume}{162}.
  \bibinfo{pages}{3648--3661}.
\newblock


\bibitem[Chen et~al\mbox{.}(2018)]%
        {chen2018neural}
\bibfield{author}{\bibinfo{person}{Ricky~TQ Chen}, \bibinfo{person}{Yulia
  Rubanova}, \bibinfo{person}{Jesse Bettencourt}, {and}
  \bibinfo{person}{David~K Duvenaud}.} \bibinfo{year}{2018}\natexlab{}.
\newblock \showarticletitle{Neural ordinary differential equations}. In
  \bibinfo{booktitle}{\emph{NeurIPS}}, Vol.~\bibinfo{volume}{31}.
\newblock


\bibitem[Cheng et~al\mbox{.}(2021)]%
        {cheng2020multiview}
\bibfield{author}{\bibinfo{person}{Jiafeng Cheng}, \bibinfo{person}{Qianqian
  Wang}, \bibinfo{person}{Zhiqiang Tao}, {and} \bibinfo{person}{Quanxue Gao}.}
  \bibinfo{year}{2021}\natexlab{}.
\newblock \showarticletitle{Multi-View Attribute Graph Convolution Networks for
  Clustering}. In \bibinfo{booktitle}{\emph{IJCAI}}.
  \bibinfo{pages}{2973--2979}.
\newblock


\bibitem[Cui et~al\mbox{.}(2020)]%
        {cui2020adaptive}
\bibfield{author}{\bibinfo{person}{Ganqu Cui}, \bibinfo{person}{Jie Zhou},
  \bibinfo{person}{Cheng Yang}, {and} \bibinfo{person}{Zhiyuan Liu}.}
  \bibinfo{year}{2020}\natexlab{}.
\newblock \showarticletitle{Adaptive graph encoder for attributed graph
  embedding}. In \bibinfo{booktitle}{\emph{KDD}}. \bibinfo{pages}{976--985}.
\newblock


\bibitem[ErdHos and R\'enyi(1959)]%
        {Erdos1959pmd}
\bibfield{author}{\bibinfo{person}{Paul ErdHos} {and} \bibinfo{person}{Alfr\'ed
  R\'enyi}.} \bibinfo{year}{1959}\natexlab{}.
\newblock \showarticletitle{On Random Graphs}.
\newblock \bibinfo{journal}{\emph{PM}}  \bibinfo{volume}{6}
  (\bibinfo{year}{1959}), \bibinfo{pages}{290--297}.
\newblock


\bibitem[Gong et~al\mbox{.}(2022)]%
        {gong2022attributed}
\bibfield{author}{\bibinfo{person}{Lei Gong}, \bibinfo{person}{Sihang Zhou},
  \bibinfo{person}{Xinwang Liu}, {and} \bibinfo{person}{Wenxuan Tu}.}
  \bibinfo{year}{2022}\natexlab{}.
\newblock \showarticletitle{Attributed graph clustering with dual redundancy
  reduction}. In \bibinfo{booktitle}{\emph{IJCAI}}.
\newblock


\bibitem[Hassani and Khasahmadi(2020)]%
        {hassani2020contrastive}
\bibfield{author}{\bibinfo{person}{Kaveh Hassani} {and}
  \bibinfo{person}{Amir~Hosein Khasahmadi}.} \bibinfo{year}{2020}\natexlab{}.
\newblock \showarticletitle{Contrastive multi-view representation learning on
  graphs}. In \bibinfo{booktitle}{\emph{ICML}}. \bibinfo{pages}{4116--4126}.
\newblock


\bibitem[He et~al\mbox{.}(2020)]%
        {he2020momentum}
\bibfield{author}{\bibinfo{person}{Kaiming He}, \bibinfo{person}{Haoqi Fan},
  \bibinfo{person}{Yuxin Wu}, \bibinfo{person}{Saining Xie}, {and}
  \bibinfo{person}{Ross Girshick}.} \bibinfo{year}{2020}\natexlab{}.
\newblock \showarticletitle{Momentum contrast for unsupervised visual
  representation learning}. In \bibinfo{booktitle}{\emph{CVPR}}.
  \bibinfo{pages}{9729--9738}.
\newblock


\bibitem[Hjelm et~al\mbox{.}(2019)]%
        {hjelm2018learning}
\bibfield{author}{\bibinfo{person}{R~Devon Hjelm}, \bibinfo{person}{Alex
  Fedorov}, \bibinfo{person}{Samuel Lavoie-Marchildon}, \bibinfo{person}{Karan
  Grewal}, \bibinfo{person}{Phil Bachman}, \bibinfo{person}{Adam Trischler},
  {and} \bibinfo{person}{Yoshua Bengio}.} \bibinfo{year}{2019}\natexlab{}.
\newblock \showarticletitle{Learning deep representations by mutual information
  estimation and maximization}. In \bibinfo{booktitle}{\emph{ICLR}}.
\newblock


\bibitem[Jin et~al\mbox{.}(2021)]%
        {ijcai2021p204}
\bibfield{author}{\bibinfo{person}{Ming Jin}, \bibinfo{person}{Yizhen Zheng},
  \bibinfo{person}{Yuan-Fang Li}, \bibinfo{person}{Chen Gong},
  \bibinfo{person}{Chuan Zhou}, {and} \bibinfo{person}{Shirui Pan}.}
  \bibinfo{year}{2021}\natexlab{}.
\newblock \showarticletitle{Multi-Scale Contrastive Siamese Networks for
  Self-Supervised Graph Representation Learning}. In
  \bibinfo{booktitle}{\emph{IJCAI}}. \bibinfo{pages}{1477--1483}.
\newblock


\bibitem[Jin et~al\mbox{.}(2022)]%
        {jin2021automated}
\bibfield{author}{\bibinfo{person}{Wei Jin}, \bibinfo{person}{Xiaorui Liu},
  \bibinfo{person}{Xiangyu Zhao}, \bibinfo{person}{Yao Ma},
  \bibinfo{person}{Neil Shah}, {and} \bibinfo{person}{Jiliang Tang}.}
  \bibinfo{year}{2022}\natexlab{}.
\newblock \showarticletitle{Automated self-supervised learning for graphs}. In
  \bibinfo{booktitle}{\emph{ICLR}}.
\newblock


\bibitem[Kingma and Ba(2015)]%
        {kingma2014adam}
\bibfield{author}{\bibinfo{person}{Diederik~P Kingma} {and}
  \bibinfo{person}{Jimmy Ba}.} \bibinfo{year}{2015}\natexlab{}.
\newblock \showarticletitle{Adam: A method for stochastic optimization}. In
  \bibinfo{booktitle}{\emph{ICLR}}.
\newblock


\bibitem[Kipf and Welling(2016)]%
        {kipf2016variational}
\bibfield{author}{\bibinfo{person}{Thomas~N Kipf} {and} \bibinfo{person}{Max
  Welling}.} \bibinfo{year}{2016}\natexlab{}.
\newblock \showarticletitle{Variational graph auto-encoders}. In
  \bibinfo{booktitle}{\emph{NeurIPS}}.
\newblock


\bibitem[Kipf and Welling(2017)]%
        {kipf2016semi}
\bibfield{author}{\bibinfo{person}{Thomas~N Kipf} {and} \bibinfo{person}{Max
  Welling}.} \bibinfo{year}{2017}\natexlab{}.
\newblock \showarticletitle{Semi-supervised classification with graph
  convolutional networks}. In \bibinfo{booktitle}{\emph{ICLR}}.
\newblock


\bibitem[Lee et~al\mbox{.}(2022)]%
        {lee2021augmentation}
\bibfield{author}{\bibinfo{person}{Namkyeong Lee}, \bibinfo{person}{Junseok
  Lee}, {and} \bibinfo{person}{Chanyoung Park}.}
  \bibinfo{year}{2022}\natexlab{}.
\newblock \showarticletitle{Augmentation-Free Self-Supervised Learning on
  Graphs}. In \bibinfo{booktitle}{\emph{AAAI}}, Vol.~\bibinfo{volume}{36}.
  \bibinfo{pages}{7372--7380}.
\newblock


\bibitem[Liu et~al\mbox{.}(2022a)]%
        {LiuAAAI2022}
\bibfield{author}{\bibinfo{person}{Chenghua Liu}, \bibinfo{person}{Zhuolin
  Liao}, \bibinfo{person}{Yixuan Ma}, {and} \bibinfo{person}{Kun Zhan}.}
  \bibinfo{year}{2022}\natexlab{a}.
\newblock \showarticletitle{Stationary diffusion state neural estimation for
  multiview clustering}. In \bibinfo{booktitle}{\emph{AAAI}},
  Vol.~\bibinfo{volume}{36}. \bibinfo{pages}{7542--7549}.
\newblock


\bibitem[Liu et~al\mbox{.}(2022b)]%
        {liu2022deep}
\bibfield{author}{\bibinfo{person}{Yue Liu}, \bibinfo{person}{Wenxuan Tu},
  \bibinfo{person}{Sihang Zhou}, \bibinfo{person}{Xinwang Liu},
  \bibinfo{person}{Linxuan Song}, \bibinfo{person}{Xihong Yang}, {and}
  \bibinfo{person}{En Zhu}.} \bibinfo{year}{2022}\natexlab{b}.
\newblock \showarticletitle{Deep graph clustering via dual correlation
  reduction}. In \bibinfo{booktitle}{\emph{AAAI}}, Vol.~\bibinfo{volume}{36}.
  \bibinfo{pages}{7603--7611}.
\newblock


\bibitem[Mrabah et~al\mbox{.}(2022)]%
        {mrabah2022rethinking}
\bibfield{author}{\bibinfo{person}{Nairouz Mrabah}, \bibinfo{person}{Mohamed
  Bouguessa}, \bibinfo{person}{Mohamed~Fawzi Touati}, {and}
  \bibinfo{person}{Riadh Ksantini}.} \bibinfo{year}{2022}\natexlab{}.
\newblock \showarticletitle{Rethinking Graph Auto-Encoder Models for Attributed
  Graph Clustering}.
\newblock \bibinfo{journal}{\emph{TKDE}} (\bibinfo{year}{2022}),
  \bibinfo{pages}{1--15}.
\newblock


\bibitem[Oord et~al\mbox{.}(2018)]%
        {oord2018representation}
\bibfield{author}{\bibinfo{person}{Aaron van~den Oord}, \bibinfo{person}{Yazhe
  Li}, {and} \bibinfo{person}{Oriol Vinyals}.} \bibinfo{year}{2018}\natexlab{}.
\newblock \showarticletitle{Representation learning with contrastive predictive
  coding}.
\newblock \bibinfo{journal}{\emph{arXiv preprint arXiv:1807.03748}}
  (\bibinfo{year}{2018}).
\newblock


\bibitem[Pan and Kang(2021)]%
        {pan2021multi}
\bibfield{author}{\bibinfo{person}{Erlin Pan} {and} \bibinfo{person}{Zhao
  Kang}.} \bibinfo{year}{2021}\natexlab{}.
\newblock \showarticletitle{Multi-view contrastive graph clustering}. In
  \bibinfo{booktitle}{\emph{NeurIPS}}, Vol.~\bibinfo{volume}{34}.
  \bibinfo{pages}{2148--2159}.
\newblock


\bibitem[Pan et~al\mbox{.}(2019)]%
        {pan2019learning}
\bibfield{author}{\bibinfo{person}{Shirui Pan}, \bibinfo{person}{Ruiqi Hu},
  \bibinfo{person}{Sai-fu Fung}, \bibinfo{person}{Guodong Long},
  \bibinfo{person}{Jing Jiang}, {and} \bibinfo{person}{Chengqi Zhang}.}
  \bibinfo{year}{2019}\natexlab{}.
\newblock \showarticletitle{Learning graph embedding with adversarial training
  methods}.
\newblock \bibinfo{journal}{\emph{TCYB}} \bibinfo{volume}{50},
  \bibinfo{number}{6} (\bibinfo{year}{2019}), \bibinfo{pages}{2475--2487}.
\newblock


\bibitem[Qiu et~al\mbox{.}(2020)]%
        {qiu2020gcc}
\bibfield{author}{\bibinfo{person}{Jiezhong Qiu}, \bibinfo{person}{Qibin Chen},
  \bibinfo{person}{Yuxiao Dong}, \bibinfo{person}{Jing Zhang},
  \bibinfo{person}{Hongxia Yang}, \bibinfo{person}{Ming Ding},
  \bibinfo{person}{Kuansan Wang}, {and} \bibinfo{person}{Jie Tang}.}
  \bibinfo{year}{2020}\natexlab{}.
\newblock \showarticletitle{Gcc: Graph contrastive coding for graph neural
  network pre-training}. In \bibinfo{booktitle}{\emph{KDD}}.
  \bibinfo{pages}{1150--1160}.
\newblock


\bibitem[Tu et~al\mbox{.}(2021)]%
        {tu2021deep}
\bibfield{author}{\bibinfo{person}{Wenxuan Tu}, \bibinfo{person}{Sihang Zhou},
  \bibinfo{person}{Xinwang Liu}, \bibinfo{person}{Xifeng Guo},
  \bibinfo{person}{Zhiping Cai}, \bibinfo{person}{En Zhu}, {and}
  \bibinfo{person}{Jieren Cheng}.} \bibinfo{year}{2021}\natexlab{}.
\newblock \showarticletitle{Deep fusion clustering network}. In
  \bibinfo{booktitle}{\emph{AAAI}}, Vol.~\bibinfo{volume}{35}.
  \bibinfo{pages}{9978--9987}.
\newblock


\bibitem[Van~der Maaten and Hinton(2008)]%
        {van2008visualizing}
\bibfield{author}{\bibinfo{person}{Laurens Van~der Maaten} {and}
  \bibinfo{person}{Geoffrey Hinton}.} \bibinfo{year}{2008}\natexlab{}.
\newblock \showarticletitle{Visualizing data using t-SNE}.
\newblock \bibinfo{journal}{\emph{JMLR}} \bibinfo{volume}{9},
  \bibinfo{number}{11} (\bibinfo{year}{2008}).
\newblock


\bibitem[Velickovic et~al\mbox{.}(2019)]%
        {velickovic2019deep}
\bibfield{author}{\bibinfo{person}{Petar Velickovic}, \bibinfo{person}{William
  Fedus}, \bibinfo{person}{William~L Hamilton}, \bibinfo{person}{Pietro
  Li{\`o}}, \bibinfo{person}{Yoshua Bengio}, {and} \bibinfo{person}{R~Devon
  Hjelm}.} \bibinfo{year}{2019}\natexlab{}.
\newblock \showarticletitle{Deep graph infomax}. In
  \bibinfo{booktitle}{\emph{ICLR}}, Vol.~\bibinfo{volume}{2}.
  \bibinfo{pages}{4}.
\newblock


\bibitem[Wang et~al\mbox{.}(2019)]%
        {wang2019attributed}
\bibfield{author}{\bibinfo{person}{Chun Wang}, \bibinfo{person}{Shirui Pan},
  \bibinfo{person}{Ruiqi Hu}, \bibinfo{person}{Guodong Long},
  \bibinfo{person}{Jing Jiang}, {and} \bibinfo{person}{Chengqi Zhang}.}
  \bibinfo{year}{2019}\natexlab{}.
\newblock \showarticletitle{Attributed graph clustering: A deep attentional
  embedding approach}. In \bibinfo{booktitle}{\emph{IJCAI}}.
  \bibinfo{pages}{3670--3676}.
\newblock


\bibitem[Wang et~al\mbox{.}(2017)]%
        {wang2017mgae}
\bibfield{author}{\bibinfo{person}{Chun Wang}, \bibinfo{person}{Shirui Pan},
  \bibinfo{person}{Guodong Long}, \bibinfo{person}{Xingquan Zhu}, {and}
  \bibinfo{person}{Jing Jiang}.} \bibinfo{year}{2017}\natexlab{}.
\newblock \showarticletitle{Mgae: Marginalized graph autoencoder for graph
  clustering}. In \bibinfo{booktitle}{\emph{CIKM}}. \bibinfo{pages}{889--898}.
\newblock


\bibitem[Xia et~al\mbox{.}(2022)]%
        {xia2022progcl}
\bibfield{author}{\bibinfo{person}{Jun Xia}, \bibinfo{person}{Lirong Wu},
  \bibinfo{person}{Ge Wang}, \bibinfo{person}{Jintao Chen}, {and}
  \bibinfo{person}{Stan~Z Li}.} \bibinfo{year}{2022}\natexlab{}.
\newblock \showarticletitle{Progcl: Rethinking hard negative mining in graph
  contrastive learning}. In \bibinfo{booktitle}{\emph{ICML}}.
  \bibinfo{pages}{24332--24346}.
\newblock


\bibitem[Yang et~al\mbox{.}(2023)]%
        {yang2023cluster}
\bibfield{author}{\bibinfo{person}{Xihong Yang}, \bibinfo{person}{Yue Liu},
  \bibinfo{person}{Sihang Zhou}, \bibinfo{person}{Siwei Wang},
  \bibinfo{person}{Wenxuan Tu}, \bibinfo{person}{Qun Zheng},
  \bibinfo{person}{Xinwang Liu}, \bibinfo{person}{Liming Fang}, {and}
  \bibinfo{person}{En Zhu}.} \bibinfo{year}{2023}\natexlab{}.
\newblock \showarticletitle{Cluster-guided Contrastive Graph Clustering
  Network}.
\newblock \bibinfo{journal}{\emph{arXiv preprint arXiv:2301.01098}}
  (\bibinfo{year}{2023}).
\newblock


\bibitem[You et~al\mbox{.}(2020)]%
        {you2020graph}
\bibfield{author}{\bibinfo{person}{Yuning You}, \bibinfo{person}{Tianlong
  Chen}, \bibinfo{person}{Yongduo Sui}, \bibinfo{person}{Ting Chen},
  \bibinfo{person}{Zhangyang Wang}, {and} \bibinfo{person}{Yang Shen}.}
  \bibinfo{year}{2020}\natexlab{}.
\newblock \showarticletitle{Graph contrastive learning with augmentations}. In
  \bibinfo{booktitle}{\emph{NeurIPS}}, Vol.~\bibinfo{volume}{33}.
  \bibinfo{pages}{5812--5823}.
\newblock


\bibitem[Zhang et~al\mbox{.}(2019)]%
        {zhang2019attributed}
\bibfield{author}{\bibinfo{person}{Xiaotong Zhang}, \bibinfo{person}{Han Liu},
  \bibinfo{person}{Qimai Li}, {and} \bibinfo{person}{Xiao-Ming Wu}.}
  \bibinfo{year}{2019}\natexlab{}.
\newblock \showarticletitle{Attributed graph clustering via adaptive graph
  convolution}. In \bibinfo{booktitle}{\emph{IJCAI}}.
  \bibinfo{pages}{4327--4333}.
\newblock


\bibitem[Zhao et~al\mbox{.}(2021)]%
        {zhao2021graph}
\bibfield{author}{\bibinfo{person}{Han Zhao}, \bibinfo{person}{Xu Yang},
  \bibinfo{person}{Zhenru Wang}, \bibinfo{person}{Erkun Yang}, {and}
  \bibinfo{person}{Cheng Deng}.} \bibinfo{year}{2021}\natexlab{}.
\newblock \showarticletitle{Graph Debiased Contrastive Learning with Joint
  Representation Clustering}. In \bibinfo{booktitle}{\emph{IJCAI}}.
  \bibinfo{pages}{3434--3440}.
\newblock


\bibitem[Zhu et~al\mbox{.}(2021a)]%
        {zhu2021contrastive}
\bibfield{author}{\bibinfo{person}{Hao Zhu}, \bibinfo{person}{Ke Sun}, {and}
  \bibinfo{person}{Peter Koniusz}.} \bibinfo{year}{2021}\natexlab{a}.
\newblock \showarticletitle{Contrastive laplacian eigenmaps}. In
  \bibinfo{booktitle}{\emph{NeurIPS}}, Vol.~\bibinfo{volume}{34}.
  \bibinfo{pages}{5682--5695}.
\newblock


\bibitem[Zhu et~al\mbox{.}(2020)]%
        {zhu2020deep}
\bibfield{author}{\bibinfo{person}{Yanqiao Zhu}, \bibinfo{person}{Yichen Xu},
  \bibinfo{person}{Feng Yu}, \bibinfo{person}{Qiang Liu}, \bibinfo{person}{Shu
  Wu}, {and} \bibinfo{person}{Liang Wang}.} \bibinfo{year}{2020}\natexlab{}.
\newblock \showarticletitle{Deep graph contrastive representation learning}. In
  \bibinfo{booktitle}{\emph{ICML Workshop on Graph Representation Learning and
  Beyond}}.
\newblock


\bibitem[Zhu et~al\mbox{.}(2021b)]%
        {zhu2021graph}
\bibfield{author}{\bibinfo{person}{Yanqiao Zhu}, \bibinfo{person}{Yichen Xu},
  \bibinfo{person}{Feng Yu}, \bibinfo{person}{Qiang Liu}, \bibinfo{person}{Shu
  Wu}, {and} \bibinfo{person}{Liang Wang}.} \bibinfo{year}{2021}\natexlab{b}.
\newblock \showarticletitle{Graph contrastive learning with adaptive
  augmentation}. In \bibinfo{booktitle}{\emph{WWW}}.
  \bibinfo{pages}{2069--2080}.
\newblock


\end{thebibliography}

%\begin{acks}
%  This work was supported by the National Natural Science Foundation of China under the Grant No. 62176108, the Natural Science Foundation of Gansu Province of China under Grant No. 20JR5RA246, and the Fundamental Research Funds for the Central Universities under the Grant No. lzujbky-2021-ct09. 
%\end{acks}

\clearpage

\appendix

\section{Experimental Details}
\subsection{Hyper-parameters settings} \label{Hyperparameters}
In this section, we list the hyperparameters utilized in our node clustering model for each of the datasets used in our experiments. The relevant parameters include $\beta$, $\gamma$, k, time, the hidden dimensions, and the number of epochs. Table~\ref{Hyper1} summarizes the hyperparameters for each dataset.
\begin{table}[!ht]
  \centering
  \caption{Hyperparameter of the node clustering.
    }
    \label{Hyper1}
  \begin{tabular}{lccccccc}
      \toprule
      Datasets  &$\beta$  &$\gamma$ &k & time &$hid\_dim$  &epochs  \\ \hline
      Cora &3 &1 &21 &100 &512 &50  \\ 
      Citeseer &7 &1 &111 &150 &512 &20 \\ 
      AMAP  &5 &1 &19  &40  &512 &20  \\
	BAT &0.7 &1 &21 &200 &64 &25 \\
	EAT &6 &1.5  &155 &15 &512 &30 \\
	CoraFull &2   &1 &73  &5 &1024 &12  \\
     \bottomrule 
  \end{tabular}
\end{table}

\subsection{Visualization analysis}
\begin{figure}[h]
\centering
	  \includegraphics[width=0.11\textwidth]{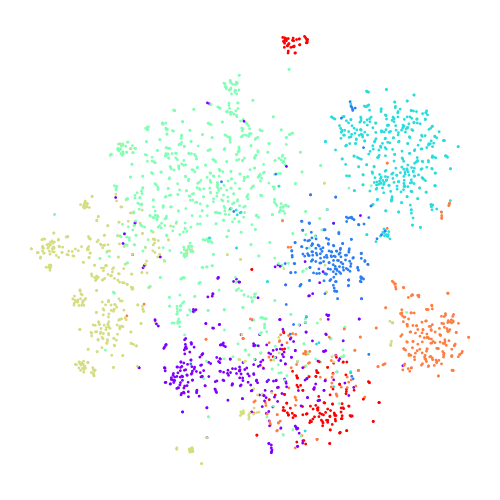}
	  \includegraphics[width=0.11\textwidth]{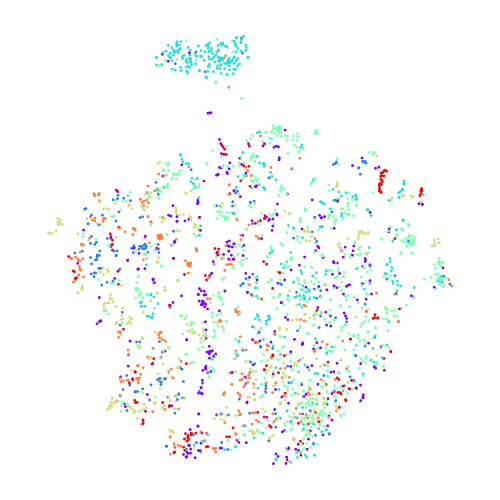}
	  \includegraphics[width=0.11\textwidth]{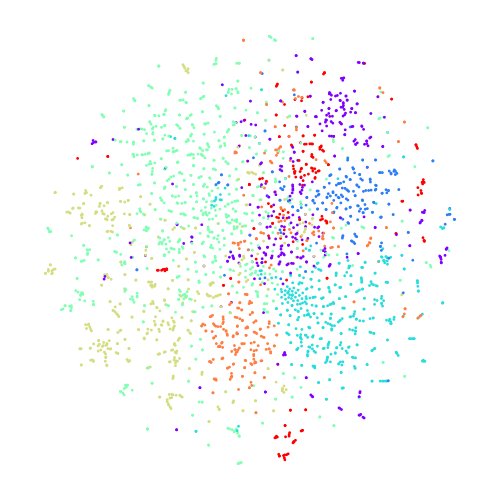}
	  \includegraphics[width=0.11\textwidth]{figures/cora_ours.png}\\
	\subcaptionbox*{MVGRL} {
	  \includegraphics[width=0.11\textwidth]{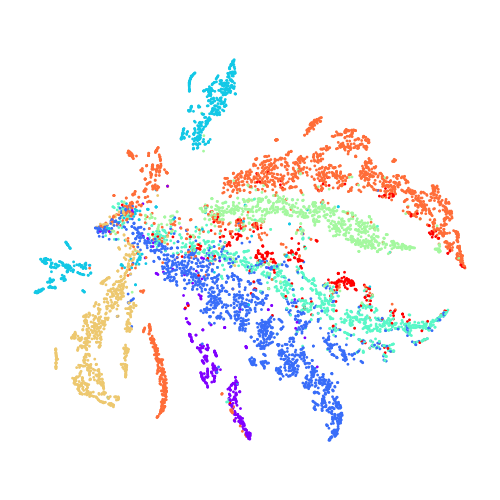}}
	\subcaptionbox*{AGC-DRR} {
	  \includegraphics[width=0.11\textwidth]{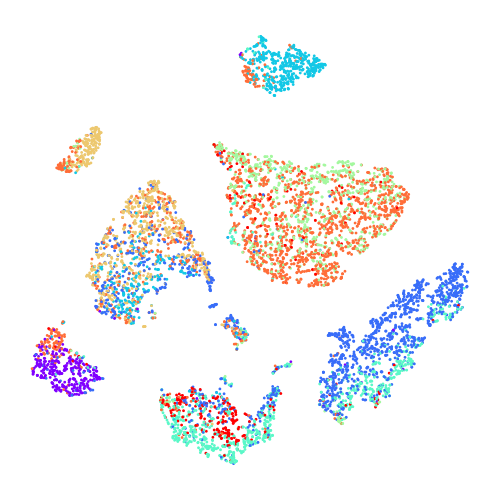}}
	\subcaptionbox*{AFGRL} {
	  \includegraphics[width=0.11\textwidth]{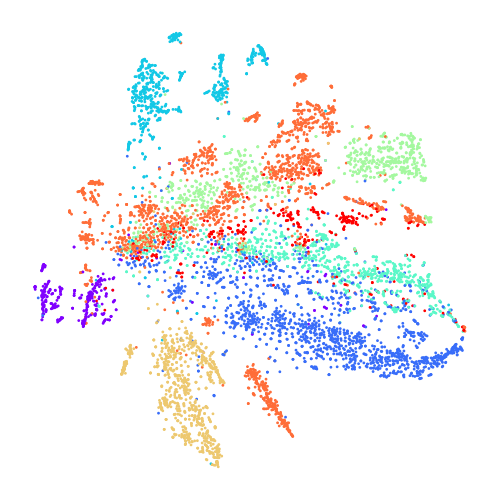}}
	\subcaptionbox*{SCGDN} {
	  \includegraphics[width=0.11\textwidth]{figures/amap_ours.png}}
        \caption{2D t-SNE visualization of six methods on two benchmark datasets. The first row and second row corresponds to Cora and AMAP datasets, respectively.}
  \label{Vis1}
\end{figure}

In this part, we visualize the distribution of the learned representation to show the superiority of SCGDN on Cora and AMAP datasets via t-SNE algorithm~\cite{van2008visualizing}. 
Three baselines and SCGDN are shown in Fig.~\ref{Vis1}, we can conclude that SCGDN better reveals the intrinsic clustering structure. 

%\begin{figure*}[hb]
%\centering
%	  \includegraphics[width=0.155\textwidth]{figures/cora_DAEGC.png}
%	  \includegraphics[width=0.155\textwidth]{figures/cora_MVGRL.png}
%	  \includegraphics[width=0.155\textwidth]{figures/cora_AGC-DRR.png}
%	  \includegraphics[width=0.155\textwidth]{figures/cora_AFGRL.png}
%	  \includegraphics[width=0.155\textwidth]{figures/cora_ProGCL.png}
%	  \includegraphics[width=0.155\textwidth]{figures/cora_ours.png}\\
%	\subcaptionbox*{DAEGC} {
%	 \includegraphics[width=0.155\textwidth]{figures/amap_DAEGC.png}}
%	\subcaptionbox*{MVGRL} {
%	  \includegraphics[width=0.155\textwidth]{figures/amap_MVGRL.png}}
%	\subcaptionbox*{AGC-DRR} {
%	  \includegraphics[width=0.155\textwidth]{figures/amap_AGC-DRR.png}}
%	\subcaptionbox*{AFGRL} {
%	  \includegraphics[width=0.155\textwidth]{figures/amap_AFGRL.png}}
%	\subcaptionbox*{ProGCL} {
%	  \includegraphics[width=0.155\textwidth]{figures/amap_ProGCL.png}}
%	\subcaptionbox*{SCGDN} {
%	  \includegraphics[width=0.155\textwidth]{figures/amap_ours.png}}
%        \caption{2D t-SNE visualization of six methods on two benchmark datasets. The first row and second row corresponds to Cora and AMAP datasets, respectively.}
%  \label{Vis1}
%\end{figure*}

\end{document}